\documentclass[manuscript]{acmart}

\AtBeginDocument{%
  \providecommand\BibTeX{{%
    \normalfont B\kern-0.5em{\scshape i\kern-0.25em b}\kern-0.8em\TeX}}}

\setcopyright{acmcopyright}
\copyrightyear{2024}
\acmYear{2024}
\acmDOI{XXXXXXX.XXXXXXX}
\acmConference[]{}{}{}
\acmPrice{15.00}
\acmISBN{978-1-4503-XXXX-X/18/06}
\acmSubmissionID{1031}

\def\shownotes{0} 
\usepackage{times}
\usepackage{epsfig}
\usepackage{graphicx}
\usepackage{amsmath}

\newcommand{\methodName}{PrivatEyes} 
\newcommand{\attackName}{DualView} 

\newcommand*{\affmark}[1][*]{\textsuperscript{#1}}
\newcommand{\share}[1]{[#1]}
\newcommand{\IU}{\textsf{IU}}
\newcommand{\OM}{\textsf{OM}}

\usepackage{enumitem}
\usepackage{stmaryrd}
\usepackage{xcolor, soul}
\usepackage{tabularx}
\usepackage{multirow}

\usepackage[capitalize]{cleveref}
\crefname{section}{Section}{Sections}
\Crefname{section}{Section}{Sections}
\Crefname{table}{Table}{Tables}
\crefname{table}{Tab.}{Tabs.}
\renewcommand{\mod}{\operatorname{mod}}

\newcommand{\Pub}{\mathsf{Pub}}
\newcommand{\Leak}{\mathsf{Leak}}

\newcommand{\Adv}{\mathcal A}

\usepackage[normalem]{ulem}
\ifnum\shownotes=1 
\newcommand\mayar[1]{\textcolor{blue}{\textbf{Mayar:} #1}}
\newcommand\andreas[1]{\textcolor{orange}{\textbf{Andreas:} #1}}
\newcommand\ralf[1]{\textcolor{cyan}{\textbf{Ralf:} #1}}
\newcommand\pr[1]{\textcolor{violet}{\textbf{Pascal:} #1}}
\newcommand\zhiming[1]{\textcolor{brown}{\textbf{Zhiming:} #1}}

\else 
\newcommand\mayar[1]{}
\newcommand\andreas[1]{}
\newcommand\ralf[1]{}
\newcommand\pr[1]{}
\newcommand\zhiming[1]{}
\fi

\begin{document}

\title[PrivatEyes]{PrivatEyes: Appearance-based Gaze Estimation Using Federated Secure Multi-Party Computation}


\author{Mayar Elfares}
\authornotemark[1]\authornotemark[2]
\email{mayar.elfares@vis.uni-stuttgart.de}
\author{Pascal Reisert}
\authornotemark[2]
\email{pascal.reisert@sec.uni-stuttgart.de}
\author{Zhiming Hu}
\authornotemark[1]\authornotemark[3]
\email{zhiming.hu@vis.uni-stuttgart.de}
\author{Wenwu Tang}
\email{}
\author{Ralf K{\"u}sters}
\authornotemark[2]
\email{ralf.kuesters@sec.uni-stuttgart.de}
\author{Andreas Bulling}
\authornotemark[1]
\email{andreas.bulling@vis.uni-stuttgart.de}
\affiliation{%
\institution{\affmark[*] Institute for Visualisation and Interactive Systems, 
  \affmark[$\dagger$] Institute of Information Security,
  \affmark[$\ddagger$] Institute for Modelling and Simulation of Biomechanical Systems,
  University of Stuttgart} 
  \country{Germany}
}

\renewcommand{\shortauthors}{Elfares et al.}

\begin{abstract}
Latest gaze estimation methods require large-scale training data but their collection and exchange pose significant privacy risks.
We propose \textit{\methodName} -- the first privacy-enhancing training approach for appearance-based gaze estimation based on federated learning (FL) and secure multi-party computation (MPC). 
\methodName~enables training 
gaze estimators on multiple local datasets across different users and server-based secure 
aggregation of the individual estimators' updates. 
\methodName~guarantees that individual gaze data remains private even if a majority of the aggregating servers is malicious.
We also introduce a new data leakage attack 
\textit{\attackName}~that shows that \methodName~limits the leakage of private training data more effectively than previous approaches.
Evaluations on the 
MPIIGaze, MPIIFaceGaze, GazeCapture, and NVGaze datasets further show that the improved privacy does not lead to a lower gaze estimation accuracy or substantially higher computational costs -- both of which are on par with its non-secure counterparts. 


\end{abstract}




\maketitle

\section{Introduction}

Starting with pioneering work by Zhang et al. \cite{zhang15_cvpr,zhang2017s}, research on appearance-based gaze estimation using deep learning has
spurred an increasing number of papers in recent years.
Much of the improvements in terms of gaze estimation accuracy can be attributed to the availability of ever-larger training datasets \cite{zhang19_pami,zhang2020eth, smith2013gaze, sugano2014learning}. 
Large eye image training data are required to capture the significant variability in eye appearances across users, tasks, and settings.
Performance could be improved significantly using data collected in the wild, e.g., on portable devices used during everyday activities \cite{zhang19_pami,bace20_chi,krafka2016eye}.
As such, it is likely that continual learning approaches will also be used in the future to collect large-scale data in the background and train personalised gaze estimators across multiple devices \cite{zhang18_chi}.

Large-scale collection and transfer of gaze data or gaze estimation models over networks, however, pose significant privacy risks, such as information leakage or misuse of gaze data copies.
With eye tracking becoming pervasive \cite{bulling2010toward, tonsen2017invisible} and integrated into an ever-increasing number of personal devices \cite{huang2016building, huang2017screen, bace2020quantification}, these privacy risks increase further. 
This is particularly critical given that gaze data contains rich personal information, such as gender \cite{sammaknejad2017gender}, identity \cite{cantoni2015gant}, personality traits \cite{hoppe2018eye}, user activities \cite{bulling2013eye, steil2015discovery, zhang2017everyday}, attentive \cite{faber2018automated, vertegaal2003attentive} and
cognitive states \cite{huang2016stress,bulling11_ubicomp}, or mental disorders \cite{hutton1984eye, holzman1974eye}.
Despite these risks, privacy has so far largely been neglected by the gaze estimation community.
One notable exception is \cite{elfares2022federated} in which the authors have proposed to increase the privacy of gaze estimators using \textit{federated learning} (FL) \cite{mcmahan2017fl}.
Their method allows training gaze estimators across a large number of clients 
without directly revealing their private data, while adapting to the heterogeneous gaze data distributions (e.g. gaze range, head pose, illumination condition,
and personal appearance).
However, it still requires each client to reveal her individually trained models (\IU) to an aggregating server and therewith remains susceptible to a large number of attacks, leaking information about the gaze data inputs as we show in this paper. 

\begin{figure}[t]
\centering
  \includegraphics[width=14cm]{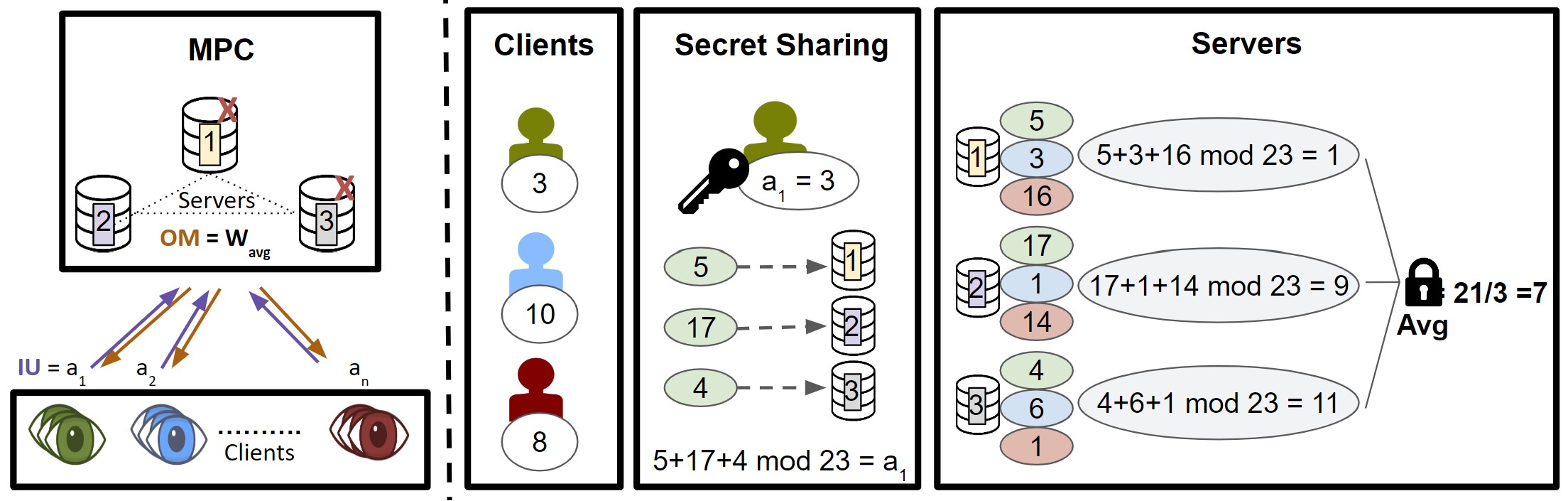}
    \caption{\methodName~combines federated learning (FL) and secure multi-party computation (MPC) for privacy-enhancing training of appearance-based gaze estimation methods (see \cref{sec:background,sub: privatEyesTrain}).  Clients $C_j$ locally train the gaze estimation model on their private data. 
    Each client $C_j$ splits her individually trained model parameters (\IU) $a_j$ into $n$ secret shares, i.e. in this example $n=3$. Each server $S_i,1\leq i,\leq n$ receives its respective share of each $a_j$, e.g. $S_1$
    receives the share $5$ of $a_1=3$, $3$ of $a_2=10$ and $16$ of $a_3=8$.
    Then each server aggregates the different shares, e.g. $S_1$ computes $5+3+16 \mod 23=1$, and sends the result (i.e. output model \OM) to \emph{all} clients. The clients compute the average $(1+9+11)/3=7$, but nothing more. \vspace{-0.35cm}
    }
\label{fig:Federated-MPC}
\end{figure}

To address these limitations, we propose \emph{\methodName} -- a novel training approach for appearance-based gaze estimation that combines FL with \textit{secure multi-party computation} (MPC) (see \autoref{fig:Federated-MPC}). In contrast to \cite{elfares2022federated}, \methodName~uses $n$ instead of only one aggregating server.
Our MPC approach then allows a gaze estimator to be jointly trained by the clients and the $n$ servers using shared secret representations of eye and face image data while keeping the data itself private to each party. 
By using MPC, it is guaranteed that no server learns the individual inputs or the individual model updates \IU~of the clients even if all-but-one server are malicious. We show that our combination of FL and MPC nevertheless only comes with a small 
computational overhead (see \cref{sec:experiments} for the exact numbers)
.
In addition, through empirical evaluation on the MPIIGaze \cite{zhang15_cvpr}, MPIIFaceGaze \cite{zhang2017s}, GazeCapture \cite{krafka2016eye}, 
and NVGaze \cite{kim2019nvgaze} datasets, we show that \methodName~ maintains an on-par gaze estimation performance as the non-secure state-of-the-art \cite{elfares2022federated}, is domain-agnostic (i.e. can be used with any deep learning-based gaze estimation model), and can scale to $\sim\! 1.5k$ clients (i.e. the size of the largest evaluated data set GazeCapture). We note that \methodName~works with \textit{any} (even larger) number of clients. 
We demonstrate the privacy advantages of our method against well-established data leakage attacks 
and our new \attackName~attack, which is able to simultaneously attack users' appearance (View1: how the user looks like) as well as their gaze distribution (View2: where the user is looking).
In summary, our work makes the following contributions:
\begin{itemize}[leftmargin=*, noitemsep]   
    \item We introduce \methodName~-- the first privacy-preserving training approach for appearance-based gaze estimation that combines federated learning and secure multi-party computation and guarantees that data collectors (servers) do not learn individual inputs by clients, even if all-but-one data collectors are malicious. 

    \item We further propose \attackName~-- a novel data leakage attack to empirically demonstrate and measure the potential privacy risks associated with gaze estimation models. 
    
    \item We implemented both \methodName~and the attack \attackName.
    Our evaluation on several gaze estimation benchmark datasets shows that \methodName~reaches the same model performance and scalability as non-secure alternatives like \cite{elfares2022federated} with negligible computational overhead. 
    \item We compare \methodName~to data centre training and federated learning \cite{elfares2022federated} w.r.t. their privacy leakage when attacked by \attackName~and other well-established data leakage attacks.
    We show that \methodName~provides significantly better privacy guarantees, e.g. for the state-of-the-art centralised FL scheme \cite{elfares2022federated} \attackName~reconstructs 15/15 participants from MPIIGaze/MPIIFaceGaze
    accurately but $0/15$ for \methodName.
    

 \end{itemize}


\section{Background}\label{sec:background}

\noindentparagraph{Federated learning} Federated learning~\cite{mcmahan2017fl} is a machine learning (ML) approach where multiple clients (e.g. gaze data owners) collaborate in solving an ML problem by training an ML model jointly (e.g. gaze estimator) on their local training dataset without sharing their data.
An aggregation server broadcasts an initial ML model to all clients.
Clients train the model on their local data samples without transferring the raw data, instead, \textit{individual updates} \IU~ (i.e. the ML model parameters) are sent to the server. The server aggregates the individual updates to a \emph{output model} \OM~which is returned to all clients for the next training round. This process is repeated until the model converges or a certain number of rounds is reached.
FL training can be performed in a centralised (client-server) or decentralised (peer-to-peer) fashion and can be classified as cross-device or cross-silo depending on whether clients are mobile devices or organisations (e.g., medical, financial, or geo-distributed data centres) that train on siloed data, respectively.
FL can be further categorised into horizontal or vertical settings according to how data is partitioned among the clients in the feature and sample spaces. In horizontal FL, clients share overlapping data features that differ in data samples while the opposite is true in vertical FL \cite{kairouz2019advances}. 

\noindentparagraph{Secure multi-party computation}\label{mpc-background} In the basic secure multi-party computation (MPC) 
setting, a set of parties $P_1,\ldots, P_n$ is considered where (i) each party has some (private) input, i.e., input that the party does not want to reveal to other parties, including external observers, and ii) the parties would like to compute a previously agreed-upon function over their private inputs without revealing any information apart from the actual output of this function. For example, the parties might want to compute their average age without revealing their respective individual age to each other. Intuitively, this can be achieved by using a completely trusted third-party (a notary): The parties could give their private input to the notary, and the notary (correctly) computes the agreed-upon function and reveals the outcome to the parties, but never leaks the individual inputs of the parties to other parties or external observers. An MPC protocol achieves exactly the same result
without actually using a notary. Instead, the MPC protocols allow the parties to exchange (a series of) specially crafted messages that allow to compute the result without ever revealing the inputs.

More formally, MPC protocols should ensure (in a mathematically rigorous way) \emph{input privacy}, i.e. that no information on the inputs (e.g. the individual age) is leaked apart from what can be deduced from the output (e.g. the average age).
Moreover, we require \emph{correctness}, i.e. every output (e.g. the average age) is correct (or the protocol aborts).
Furthermore, we require that both input privacy and correctness still hold if all but one of the computing parties act \emph{maliciously}, e.g. by colluding with each other or by deviating from the protocol. 
This means that a single honest MPC party still guarantees that inputs remain private and that the result is correct even if all other parties act maliciously. Note that even if the malicious parties force a protocol abort, e.g. by simple denial of service, the privacy of inputs is guaranteed.
MPC protocols that ensure the aforementioned security properties are called \emph{actively secure} against a majority of up to $n-1$ malicious parties.
The currently best known MPC protocols that provide this security standard are SPDZ \cite{damgaard2012spdz} and improvements thereof like Overdrive LowGear \cite{keller2018overdrive}. 


We employ these state-of-the-art protocols in a client/server MPC model.
That is, we distinguish between input parties (clients), who provide the inputs on which the function is evaluated, and compute parties (servers), who carry out the MPC protocol and provide the clients with the output of the function evaluation. 
Then as long as one server remains honest, the inputs of all clients remain private and the result is correct (or the protocol aborts). 
The client/server model is particularity well-suited for a setup with (possibly resource-limited) clients and a few (more powerful) servers as it is common in our gaze applications \cite{kairouz2019advances, zhang15_cvpr, elfares2022federated}.

\section{Related Work}\label{sec:relatedwork}

\noindentparagraph{Gaze estimation} Gaze estimation methods can generally be categorised as model-based or appearance-based \cite{hansen2009beholder}.
While model-based methods estimate gaze direction from infrared corneal reflections \cite{hennessey2006single, morimoto2002detecting},
or geometric eye shape \cite{chen20083d, valenti2011combining}),
appearance-based methods directly regress a 2D or 3D gaze direction from eye images recorded using an off-the-shelf RGB camera \cite{baluja1993non,liang2013appearnce,Choi2013appearnce}.
Appearance-based methods, particularly those based on deep learning, were shown to be more robust to varying lighting conditions, gaze ranges, and low image resolutions than their model-based counterparts \cite{zhang15_cvpr,zhang2017s,biswas2021appearance}.
However, while gaze estimators can also be trained from synthesised images \cite{wood16_etra,sugano2014learning,wood16_eccv}, effective training typically requires a large number of real-world training images \cite{krafka2016eye,zhang19_pami,zhang2020eth}.
Despite an increasing number of works and significant advances in appearance-based gaze estimation methods in recent years, the privacy threats posed by the requirement for large-scale image datasets collected of individuals remain under-explored in the community. We address these privacy threats by presenting the first privacy-preserving gaze estimation scheme \methodName~in \cref{sec:method}.

\noindentparagraph{Privacy-preserving machine learning}
Initially, privacy-preserving machine learning either implied low efficiency \cite{gilad2016cryptonets, burkhalter2021rofl, chowdhury2021eiffel}, weak privacy guarantees \cite{mohassel2017secureml, liu2017minionn,rouhani2018deepsecure, mohassel2018aby3}, 
or was restricted to specific setups \cite{mohassel2017secureml, liu2017minionn,  galli2022machine}. 
While the efficiency of distributed ML training has improved with FL, privacy issues endured due to the lack of provable privacy guarantees \cite{liu2022threats}. 
First secure aggregation protocols based on MPC \cite{bonawitz2017practical} address these issues but either come with a low efficiency \cite{bonawitz2017practical, dong2021flod, chowdhury2021eiffel} or restrict to a weak security setup, i.e. assume that all servers always follow the MPC protocol
\cite{fereidooni2021safelearn, dong2021flod, nguyen2022flame, rathee2023elsa}. In addition, they have not been adapted to the requirements of gaze estimation (e.g. the prior eye/face knowledge, the in-the-wild data heterogeneity, and the computational performance). 
We are the first to practically combine FL with state-of-the-art MPC protocols like Overdrive LowGear \cite{keller2018overdrive} 
in the context of appearance-based gaze estimation in order to achieve active security against a dishonest majority of servers and allow scalability to a large number of clients. 

\noindentparagraph{Privacy-enhancing technologies for gaze}
Despite mounting concerns, only few previous works have studied privacy-enhancing technologies for gaze, most of which have focused on gaze behaviour analysis.
A common approach is to use differential privacy (DP), i.e., the idea of adding noise according to an $\epsilon$ parameter \cite{dwork2014algorithmic,steil2019privacy}.
This approach, however, often comes at the cost of reduced data utility \cite{li2021kaleido,bozkir2021diffrential,liu2019differential}.
Others have proposed to add noise in a dataset-dependent manner to increase utility \cite{bozkir2021diffrential, steil2019privacy}, which leads to very weak privacy guarantees \cite{nissim2007smooth, bozkir2023eye}. 
Further DP optimizations, e.g. as in \cite{kifer2011no}, are possible, but come with a decrease in the model performance.
In stark contrast, privacy for gaze estimation has so far largely been neglected except for \cite{elfares2022federated, bozkir2020privacy}. In \cite{bozkir2020privacy}, a function-specific (support vector regressor) method is proposed using randomized encodings. This method is limited by two data owners and does not protect against malicious adversaries.
Our work, therefore aims to fill these gaps and builds on the adaptive federated learning approach proposed in \cite{elfares2022federated}. Adaptive FL comes with many favourable properties, e.g. it scales well in the number of clients, can be applied to any ML model and converges quickly even for a large data heterogeneity across the clients. However, we show in this paper that it remains susceptible to various data leakage attacks. Our new training schemes \methodName~solves this privacy issue while maintaining the same model performance and comparable computational cost as \cite{elfares2022federated}.

\noindentparagraph{Adversarial attacks}
ML models are susceptible to a large number of vulnerabilities and attacks. The trained models can leak information about the private input data through their black-box outputs (e.g. predictions) or their white-box parameters \cite{zhang2020secret}. 
As no prior work has investigated adversarial attacks on gaze estimation models, in this paper, we aim (i) to analyse these attacks on gaze data and its training process, (ii) to quantify the amount of information leakage and (iii) to prevent (or at least minimise the effectiveness of) such attacks through protocols that provide formal
security guarantees, i.e. which do not only protect against a certain attack but provide security against any (realistic) adversary and attack.
Prior works in ML investigated different attacks in isolation, e.g. model-inversion attacks  \cite{fredrikson2015model, he2019model, wu2016methodology} (i.e. reconstructing the training data) or inference attacks  \cite{shokri2017membership, bernau2019assessing, li2020membership} (i.e. inferring private information). Other approaches \cite{hitaj2017deep, zhao2020idlg, geiping2020inverting} were specific to the FL setup.
Given the various different gaze estimation training approaches, we instead construct a new attack \attackName~that allows us to attack different schemes like data centre training, adaptive FL \cite{elfares2022federated} or our own approach \methodName.
Unlike (adversarial) model-inversion attacks \cite{fredrikson2015model, zhang2020secret, hitaj2017deep} that reconstruct images that maximally activate the target network, \attackName~does not only aim to synthesize realistic features but also tries to consistently associate the reconstructed images with the appearance and the gaze features of the training set.
\attackName~is further optimized for the regression task of gaze estimation and therewith differs from classical inference attacks 
like \cite{shokri2017membership, zhao2020idlg, salem2020updates}
which were so far only studied for classification tasks.
We use \attackName~among other techniques to show the vulnerabilities of FL training. For FL applications (without MPC) outside of gaze estimation, similar results have been achieved by Geiping et al. \cite{geiping2020inverting}.

\section{Method}\label{sec:method}

Training accurate and generalisable gaze estimation models requires a large number of training images to handle the large variability in in-the-wild eye and face appearances. 
A common solution is to train gaze estimators collaboratively using gaze data from different owners (a.k.a. clients) but this raises concerns with the highly privacy-sensitive gaze data. 
Missing protection of this data prevents gaze data owners from taking part in the training process and thereby decreases the model generalisation performance.
We address this problem by introducing in a new gaze estimation training approach \textit{\methodName} that provides strong security guarantees without hampering training efficiency or gaze estimation performance.
We further construct a new gaze-specific attack that allows us to quantitatively compare the privacy properties of different gaze estimation training schemes like adaptive FL \cite{elfares2022federated} or our new scheme \methodName.

\subsection{ \methodName: Gaze Estimation Training} \label{sub: privatEyesTrain}
FL has recently been introduced as a promising approach for training gaze estimation models \cite{elfares2022federated}. 
Following \cite{elfares2022federated}, we focus on a cross-device centralised horizontal FL approach, with its support for scalable data distributions and a large number of devices, hence matching the requirements of large-scale gaze data collection on multiple devices in the wild. 
However, centralised FL comes with a major privacy issue: It involves a single central server that aggregates local updates received from all clients. 
This results in the server having access to a significant amount of personal information that could be used to reconstruct (parts of) the original training data. 
We address this issue in \methodName~by replacing the single server with multiple ones (with secret sharing) as shown in \autoref{fig:Federated-MPC}.
This allows us to avoid the leakage of individual client updates and thereby significantly reduce the attack surface. 

\noindentparagraph{The training process}
Centralised FL with a single server (e.g. \cite{elfares2022federated}) works as follows: 
For each of $t$ communication rounds
the server selects a random cohort $C$ of the $N$ available clients and
broadcasts a gaze estimation model
to these clients
with the corresponding hyper-parameters, weights, biases, number of rounds, and number of local epochs.
In the first round ($k=1$) the model sent is some initial model (e.g. a model pre-trained on public data) and in all following rounds $(1<k\leq t)$ it is the output model (\OM$_{k-1}$) of the previous round.
Once a client $C_j\in C$ receives the model (represented by some weight vector $w$) from the server,
it starts locally training a new model \IU$_{j,k}$ on her private data $D_{j,k}$. It outputs this individually trained model (represented by some weight vector $a_j$) to the server. 
The server aggregates the individual updates \IU$_{j,k}$ to get a new output model $\OM_{k}$ represented by $\frac{1}{|C|}\sum_{j\in J}a_j$.
The process continues until a final output model $\OM_{t}$ is reached and then published to all clients.

\noindentparagraph{Secret sharing}
In contrast, in \methodName, as shown in \cref{fig:Federated-MPC}, a client $C_j$ no longer sends its update $a_j$ in plain to an aggregating server but instead encrypts it as a \emph{secret sharing} for $n$ servers. 
That is, for each of the $n$ servers $S_i$, a random number $\share{a_j}_i$ ($1\leq i \leq n$) is selected such that $a_j=\sum^n_{i=1}\share{a_j}_i$ holds;
$\share{a_j}_i$ belongs to a previously agreed finite field, such as $\mathbb Z_q ={0,1,...,q-1}$,
for some prime number $q$ (e.g. a common size is  $q\approx 2^{127}$). 
The client then sends $\share{a_j}_i$ to $S_i$, i.e. each server gets only one share of the secret update $a_j$.
As long as at least one server, say $S_2$, is honest and does not reveal its share $\share{a_j}_2$, MPC guarantees that the other servers cannot gain \emph{any} information about $a_j$. 
E.g., as shown in \cref{fig:Federated-MPC}, the client $C_1$ wants to share $a_1=3$ as an element of a finite field $\mathbb Z_{23}$ 
with three servers $S_1,S_2,S_3$.
Then $C_1$ chooses (arbitrarily) three numbers $\share{a_1}_1, \share{a_1}_2, \share{a_1}_3$ such that $(\share{a_1}_1+\share{a_1}_2+\share{a_1}_3)\mod 23=a_1=3$. For example,
$\share{a_1}_1=5, \share{a_1}_2=17$ and $\share{a_1}_3=4$ with $(5+17+4) \mod 23= 26 \mod 23 = 3$. 
Even if $S_1$ and $S_3$ collude and exchange their shares, the unknown share of $S_2$ makes all possible values of $a_j\in \mathbb Z_{23}$ equally likely from the perspective of $S_1$ or $S_3$, i.e. no information on the actual $a_j$ is leaked.
\newline
Once each server $S_i$ has received a share $\share{a_j}_i$ from each client $C_j$, the servers start an MPC protocol
to compute a \emph{sharing} of the new global model represented by a weight vector $w$.
At the end of the MPC computation, each server $S_i$ has a share $\share{w}_i$. 
As before, if only one server is honest, the MPC protocol guarantees that no server gets any information on $w$ and that $w$ has been computed correctly. 
Otherwise, the protocol aborts. Finally, all servers return their shares $\share{w}_i$ to all clients $C_j$ and each $C_j$ can locally reconstruct $w=\sum^n_{i=1}\share{w}_i$ by simply adding up the shares. 
\newline
Clients provide their inputs by following the protocol of Damgard et al. \cite{damgaard2012spdz} (simplified above).
It guarantees along with checks carried out in the MPC protocol that servers are forced to use the inputs (the shares) given to them by the clients when performing the MPC protocol.
I.e., even malicious servers cannot change the client inputs. Hence, if an output is produced by the MPC computation, this output is (mathematically provably) guaranteed to correspond to the inputs provided by the clients; otherwise, no output is produced.
For an efficient aggregation of the individual model updates, the MPC protocol evaluates an adaptive optimisation protocol \cite{reddi2020adaptive} that has been shown to adapt the model updates to the clients’ heterogeneous gaze data distribution \cite{elfares2022federated} while only relying on low-computational operations on secret values for efficient privacy-preserving training (a main challenge in MPC-based protocols). 
Following the classical FL approach, the FL training is repeated until a certain number of rounds is reached and the final \emph{output model} has been generated.
For more details on the MPC protocol, formal security and privacy guarantees, and security proofs please refer to \cite{damgaard2012spdz,keller2018overdrive,damgaard2016benchmarking}. 

\noindentparagraph{Security guarantees of \methodName}
\methodName~provides security guarantees even if only one server (out of $n$ servers) is honest, i.e., follows the protocol and does not collude with other servers or clients:

\begin{enumerate}[leftmargin=*, noitemsep]
    \item If servers do not use the input provided to them by (honest) clients the protocol aborts. 
    \item If servers deviate from the prescribed protocol the protocol aborts. 
    \item If an output is produced by the protocol, then this is guaranteed to be the correct global model, i.e., the gaze estimation model that would have been obtained if all servers were honest and used the inputs provided to them by the clients. 
    \item A server gains no information from the gaze data of a (honest) client beyond what is available publicly. 
\end{enumerate}

\subsection{\attackName: Gaze-specific Attack}\label{attack}

Gaze estimation models are susceptible to a large number of vulnerabilities and attacks (c.f. \cref{sec:relatedwork}). Nonetheless, such attacks have never been studied in the gaze community. 
In this work, we present a new gaze-specific attack,
which we call \attackName,  to evaluate the amount of information leakage from gaze estimation models and 
their training process. 

\begin{figure}[t]
\centering
  \includegraphics[width=14cm]{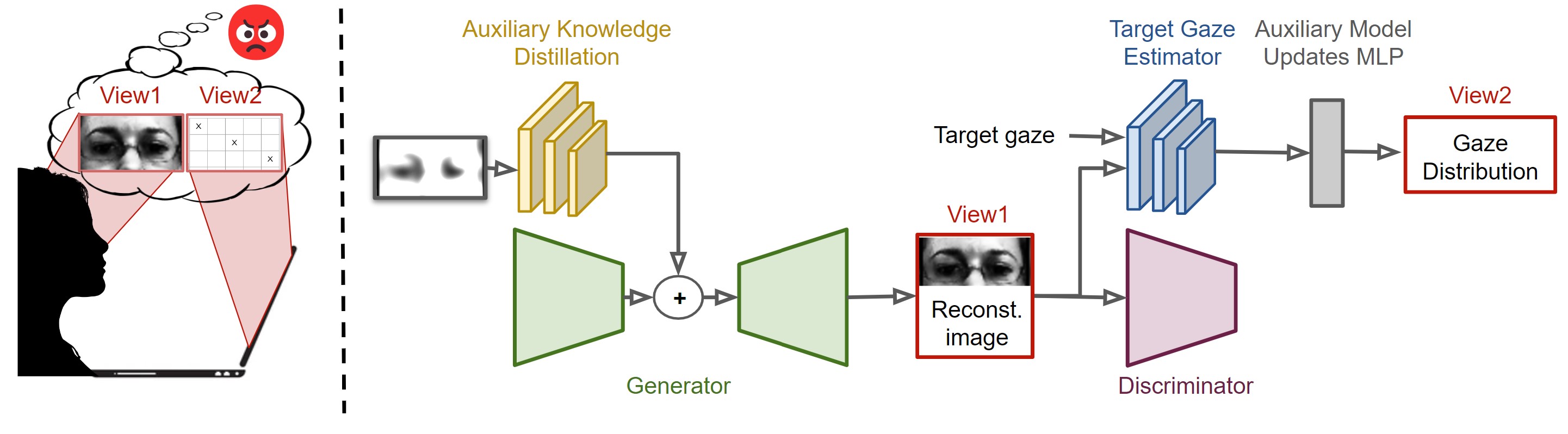}
    \caption{\attackName~ demonstrates the amount of gaze-specific information leakage by reconstructing the user’s appearance (View1: how the user looks like) as well as inferring the corresponding gaze distribution (View2: where the user is looking) of the private dataset used to train a target gaze estimation model. \vspace{-0.4cm}} 
\label{fig:Attack}
\end{figure}

\noindentparagraph{Threat Model}
We assume an adversary $\Adv$ that can use all available knowledge to determine properties of the private training data sets used by the clients. 
The knowledge of $\Adv$ contains information $\Pub$ that is publicly available, e.g., the final output model
or the model architecture.
But it can also contain leaked knowledge $\Leak$ that the adversary gathered from the training process, e.g. by colluding with some of the servers.
In particular, depending on the actual training scheme 
and the number of dishonest parties, the knowledge available to $\Adv$ differs. 
For example, in centralised FL an adversary that colludes with the aggregating server 
sees all individual model updates \IU~and can use these updates in his attack. In contrast, in \methodName~, due to the use of MPC (c.f. \cref{fig:Federated-MPC}), 
an adversary that colludes with one (or up to $n-1$) servers does not get any information on the individual updates \IU~and can, therefore, not use this information in the attack.

\noindentparagraph{Goal of the attack} For our attack, we assume the adversary $\Adv$ wants to reconstruct 
private training data 
(given his available information from $\Pub$ and $\Leak$) or some property of the training data (e.g. its gaze distribution). We then determine the success of $\Adv$ in \cref{sub:DualView} by measuring how much his reconstruction differs from the original training data (w.r.t. the ground truth).
Naturally, if a gaze estimation training scheme like centralised FL leaks information, then this information leakage likely contains gaze-related information. Unfortunately, this is not the only information one can usually deduce from the leaked data, e.g. from a model update. For example, a model update usually also contains appearance-related information (although appearance was not the main objective of the training). 
Hence, in order to get a more complete picture of the overall information leakage of a gaze estimation training scheme like centralised FL or \methodName, we need an attack 
that can successfully deduce information about a user's appearance (i.e. 
view1: how the user looks like) as well as the respective gaze angle distribution (i.e. view2: where the user is looking) in the respective training set. Our attack \attackName~(cf. \cref{fig:Attack}) addresses these two objectives.
\noindentparagraph{Our attack \attackName} Technically, \attackName~consists of a generative adversarial network (GAN) that trains a generator $G$ and a discriminator $D$, where the generator takes a random latent vector as input and generates face/eye images that look (superficially) authentic to human observers with realistic characteristics. The discriminator evaluates the reconstructed images by distinguishing the reconstructed images from the true data distribution. The generator takes a random input (sampled from a latent space) as a seed and is trained until it succeeds at fooling the discriminator via independent backpropagation with a training loss composed of the Wasserstein loss and a diversity loss \cite{yang2019diversity} (to generate a large output distribution for our domain-agnostic reconstructed images).
As a training dataset, we use a high-quality reference dataset which is publicly available, namely GazeCapture or LPW datasets, for face and eye images respectively.  We choose these datasets since they come with a large number of participants, e.g. 1,474 in the case of GazeCapture, and therefore result in a generator that can reproduce realistic reconstructions of training sets.



\noindent Note that so far our attack only used public information, i.e. the publicly available data sets GazeCapture and LPW. In particular, \attackName~generates the GAN independently of the specific training scheme and its leakage $\Leak$. 
We next want to explain how \attackName~uses the additional knowledge $\Leak$ 
gained from the gaze estimation training (e.g. by corrupting servers or clients) to reconstruct the training data sets.
Generally, the knowledge in $\Leak$ will contain some model updates, e.g. the individual model updates revealed in centralised FL. For the exact form of $\Leak$ in the different training schemes we refer to \cref{sub:DualView}. 
Now \attackName~checks how accurately such a model from $\Leak$ can determine the gaze angle on generated images by $G$. 
This leads to a loss function $L_{\text{gaze}}$. Similarly, \attackName~determines a loss $L_{\text{prior}}$ (the discriminator loss that penalises inauthentic reconstructed images) and an appearance loss $L_{\text{appearance}}$ (a cross-entropy loss). 
Overall he gets a total loss function\footnote{Weighting constants are selected according to a hyperparameter search. E.g. for attacking full-faces, $\alpha=10$, $\beta=6$, and $\gamma=4$.}:
$$ L_{\text{total}} = \alpha L_\text{prior} + \beta L_\text{gaze} + \gamma L_\text{appearance} :\alpha, \beta, \text{ and } \gamma \text{ are weighting constants.}$$
The gaze loss is then fed to a simple multi-layer perception (MLP)-based autoencoder that regresses it to the corresponding gaze direction distribution. In addition, a kernel density estimation (KDE) function derives a probability density function to compute a Kullback-Leibler (KL) divergence in order to evaluate the predicted gaze distribution.

\noindent Now generally, $\Leak$ contains more than a single model, e.g. it contains all individual model updates \IU~from all rounds in centralised FL. 
We introduce these different model updates in $\Leak$ iteratively.
Namely, \attackName~first optimizes w.r.t. the first round model update.
The resulting reconstructed images are then (re-)introduced as auxiliary knowledge following \cite{zhang2020secret}, i.e. after passing them through an auxiliary knowledge distillation network with dilated convolution layers that insert gaps between the kernel elements (i.e. pixel skipping) to cover a larger receptive field and therewith provide a generic eye/full-face skeleton to the generator. Similarly, all available model updates in $\Leak$ (and the final output model in $\Pub$) are used by the MLP-based autoencoder as auxiliary knowledge.

\noindent Finally, the adversary receives a reconstruction of the private training set that agrees best with his available knowledge from $\Pub$ and $\Leak$. We remark again that the quality of this reconstruction of course depends on $\Leak$ and that a \attackName~instance which can use all individual model updates and all round-wise output models (i.e. in the case of centralised FL) will generally produce a more accurate reconstruction than a \attackName~instance which can only use the output models without individual updates (i.e. in the case of \methodName). 

\section{Experiments}\label{sec:experiments}


To evaluate our method, we use several different datasets to cover the different gaze estimation setups and to prove the advantage of \methodName~
and our privacy guarantees in real-world scenarios. The datasets cover different appearances (ethnicities, genders, glasses, and make-up), illumination conditions (indoor and outdoor), gaze distributions, head poses, modalities (images and videos with eyes or full-faces), and recording setups (remote vs. near-eye).

We mainly conducted experiments on the MPIIGaze \cite{zhang15_cvpr}, MPIIFaceGaze \cite{zhang2017s}, and NVGaze \cite{kim2019nvgaze} datasets.
The MPIIGaze and MPIIFaceGaze datasets contain 213,659 eye and full-face images, collected in the wild from 15 participants over the course of several months.
We used the gaze estimation models originally proposed in both works and trained them using \methodName.
The models take eye/full-face images as input and regress them to gaze directions in normalised space.
We conducted further experiments for near-eye gaze estimation on the NVGaze \cite{kim2019nvgaze} dataset, with 32 participants recorded by a near-eye infrared camera, along with their proposed gaze estimation models. 
Furthermore, we used GazeCapture \cite{krafka2016eye}, the gaze estimation dataset with the largest number of participants (1,474), (1) along with its corresponding iTracker CNN, to further analyse the scalability of \methodName~and (2) to train our \attackName~attack for full-face inputs. Similarly, the LPW \cite{tonsen2016lpw} dataset, which contains videos of 22 participants recorded by a head-mounted eye tracker was used to train the \attackName~attack for eye inputs.
We used the MP-SPDZ \cite{keller2020mpspdz} framework with Overdrive LowGear \cite{keller2018overdrive} (the state-of-the-art maliciously secure MPC protocol for a small number of servers) to implement \methodName~
and the ML-Doctor \cite{liu2022mldoc} framework to also perform standard ML attacks (c.f. \cref{sub:attacks})
in addition to our new attack \attackName. 

\subsection{Baseline methods} \label{sub:baselines}
We compared our training approach \methodName~(cf. \cref{sub: privatEyesTrain}) with three baselines and evaluated them in terms of gaze estimation error, robustness to gaze data leakage attacks, client-server communication, and computational complexity:
\begin{itemize}[leftmargin=*, noitemsep]    
   \item\textbf{Data centre training:}
   In this approach, the clients' datasets are collected in one central server storage. The model is directly trained on all samples and, hence, data-sharing concerns arise.
  \item\textbf{Adaptive FL:} This is the state-of-the-art centralised federated learning-based gaze estimation training proposed in \cite{elfares2022federated} in which the individual model updates are treated as pseudo-random gradients for aggregation.

  \item\textbf{Generic MPC:} The state-of-the-art actively secure MPC approach (without FL) from \cite{keller2018overdrive} where the clients provide the inputs in a secret-shared form and the servers run the whole training as an MPC protocol (cf. \cref{mpc-background}). In particular, this approach uses only MPC and not federated training like \methodName.

\end{itemize}


\subsection{Gaze estimation performance} \label{sub:results}\label{sub:scalability}
Gaze estimation performance was calculated as the mean angular error between the predicted and ground-truth gaze directions. 
As shown in \cref{tab:mae}, for data centre training, the gaze estimation model yielded the best performance on all datasets. This is expected given that this training approach can have direct access to original images. Of course, privacy in the data centre training relies on the trustworthiness of the central server---if the server is dishonest, \emph{all} input data gets leaked (c.f. \cref{sub: privatEyesTrain}). 
For adaptive FL and \methodName, we used the same number of rounds, epochs, and hyper-parameters as in \cite{elfares2022federated}.
As expected (c.f. \cref{sub: privatEyesTrain}), the performance of adaptive FL and \methodName~ is identical.
In addition, we investigated how the performance loss between data centre on the one side and adaptive FL and \methodName~on the other side varies from client to client. For example, \cref{fig:individual} shows a random sample of $10$ clients with the respective errors.
In general the gaze estimation fairness (i.e. the difference between the minimum and maximum gaze angular error 
across all clients) is $1.3^\circ, 1.5^\circ, 1.0^\circ$, and $1.5^\circ$ for MPIIGaze, MPIIFaceGaze, NVGaze, and GazeCapture respectively. 
Finally, we were not able to run the full generic MPC benchmark for the performance, since these protocols are far too inefficient to handle realistic data sets and realistic numbers of clients (c.f. \cref{sub:efficency})\footnote{The main reason for this lack of efficiency is that operations common in ML, e.g. comparisons, polynomial evaluations and generally floating point operations, are not naturally supported by the MPC protocols, which are optimised for arithmetic operations over finite fields. This leads to costly translations and often comes with a loss in precision and therefore a loss in the model performance. 
However, theoretical considerations ensure that generic MPC will perform as good as data centre training while runtime was estimated by only running the main operations and generalising them to the entire training.}

Our evaluation in \cref{tab:mae} also indicates how our approach \methodName~scales with the number of clients and the size of the training sets.
From the four data sets in \cref{tab:mae}, MPIIGaze, MPIIFaceGaze, and NVGaze datasets represent data sets with a small number of participants (15 to 32), while GazeCapture covers a large number of participants (1,474) and therewith follows classical cross-device federated training conventions (which usually come with $\geq 100$  clients \cite{kairouz2019advances}). 
In line with \cite{kairouz2019advances},
the performance gap between data centre training/generic MPC and \methodName/adaptive FL becomes smaller with a larger number of clients and larger datasets.

\subsection{\attackName~performance}\label{sub:DualView}

In this section, we want to compare different training schemes, i.e. data centre training, adaptive FL, \methodName~and generic MPC, w.r.t. their privacy leakage. For the two federated approaches, we consider a training with $t=10$ rounds. To simplify the evaluation we further assume that $C=\{1,\ldots,N\}$, i.e. that all clients are chosen in each round.\pr{maybe a remark about different ratios.}
\begin{table}[t]
    \centering

    \begin{tabularx}{1\textwidth} 
    { 
       >{\raggedright\arraybackslash}X 
       >{\centering\arraybackslash}X
       >{\centering\arraybackslash}X
       >{\centering\arraybackslash}X
      }
     \toprule
     & \mbox{\textbf{Data Centre/Generic MPC}} & \textbf{Adaptive FL} & \textbf{PrivatEyes} \\
     \midrule
     \textbf{MPIIGaze} & 
     $6.3^\circ$  &  $\mathbf{7.6^\circ}$ & $\mathbf{7.6^\circ}$ \\
     \midrule
     \textbf{MPIIFaceGaze} & $6.2^\circ$ & $\mathbf{7.4^\circ}$ & $\mathbf{7.4^\circ}$  \\
     \midrule
     \textbf{NVGaze} & $0.8^\circ$ & $\mathbf{2.1^\circ}$ & $\mathbf{2.1^\circ}$  \\
     \midrule
     \textbf{GazeCapture} & $4.0^\circ$ & $\mathbf{4.7^\circ}$ & $\mathbf{4.7^\circ}$  \\
     \bottomrule
    \end{tabularx}
    
    \caption{Mean angular error for different gaze estimation datasets.\vspace{-0.5cm}}
    \label{tab:mae}
\end{table}

We assume that at least one and at most $n-1$ servers (if $n\geq 2$) are corrupted.
Hence, the data centre training is completely insecure and the one corrupted central server leaks all private training data. 
For adaptive FL, the corrupted server leaks all individual model updates and all output models in all rounds, i.e. $\Leak_{adaptive~FL}=\{\IU_{j,k},\OM_k:1\leq k \leq n\}$.
Furthermore, we assume that at least one client is corrupted. Recall from \cref{sub: privatEyesTrain} that in \methodName~each client receives the round-wise output models and hence a corrupted client leaks these output models to the adversary.
However, the individual models (of an honest party) are only send in shared form in \methodName~and therefore no corrupted server or client can deduce information about them.
Thus, in \methodName, we have $\Leak_{\methodName}=\{\OM_k:1\leq k \leq t\}$.
Finally, in generic MPC, no information is leaked apart from public information, e.g. the final output model $\OM_t$.

\noindent We then run three instances of \attackName: (i) against adaptive FL, where \attackName~uses $\Leak=\Leak_{adaptive~FL}$; (ii) against \methodName~, where $\Leak=\Leak_{\methodName}$ contains only all roundwise output model updates; (iii) against generic MPC, where \attackName~receive no non-public information, i.e. $\Leak_{\operatorname{MPC}}=\{\}$ and \attackName~stops once he used all publicly available information for the reconstruction.
In each of the three cases \attackName~outputs a reconstruction (i.e. View1 and View2) of the training set of each client at each round. We then evaluate the reconstructions in terms of:

\begin{itemize}[leftmargin = *]
    \item \textbf{Appearance similarity:} A user study (N = 60 respondents) was conducted to see if users could correctly map the reconstructed images to the corresponding participant.\footnote{A re-identification or image similarity neural network could serve as a metric for appearance similarity as well. However, in this work, we opted for the user study to avoid bias.}
    Additionally, the participants were asked to rate the similarity of a reconstruction with a qualitative visual score.
    See \cref{app:study} for more details.  

    \item \textbf{Pixel-wise similarity:} A Peak Signal-to-Noise Ratio (PSNR) was used to quantify the fluctuation between the original (private) image and the corresponding reconstructed image. Higher PSNR scores indicate higher similarities. 

    \item \textbf{Gaze direction similarity:} A mean angular error (MAE) was calculated to capture the gaze direction similarity between the ground truth and the reconstructed images.

    \item \textbf{Gaze distribution similarity:} A KL-divergence is used to calculate the statistical distance between the ground truth and the inferred gaze probability distributions of all images used to train a target model. A value of 0 indicates identical quantities of information (i.e. identical gaze directions).
    
\end{itemize}

\begin{figure}[t]
\centering
  \includegraphics[width=12cm]{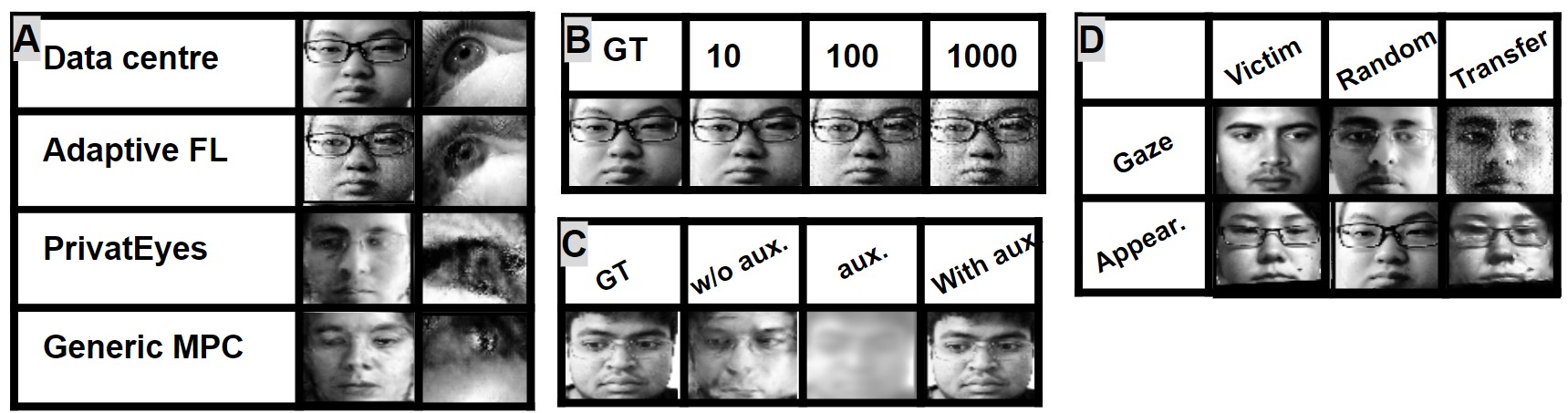}
    \caption{
    Sample reconstruction results. (A) Reconstructed face/eye images of the same participant for the different baselines (Data centre is the ground truth due to the direct access to the raw data). (B) Effect of the number of training samples per round. (C) Using previous reconstructions as auxiliary knowledge against adaptive FL training affects the reconstruction. (D) Effect of the attack loss function, e.g. transferring the gaze of a victim to a random face (1st row) and transferring the appearance of a victim to a random gaze (2nd row) in adaptive FL.
 \vspace{-0.3cm}   }
\label{fig:reconstructions}
\end{figure}
Our results for the MPIIFaceGaze dataset are included in \cref{tab:attack} and \cref{fig:reconstructions}. Some additional results are contained in \cref{sec:additional} for the remaining datasets. Note that the same performance behaviour can be seen across all datasets (c.f. \cref{sec:additional}).
As expected our evaluation shows that more leaked information naturally leads to better reconstructions of the training set, i.e. the data leakage increases from generic MPC and \methodName~to adaptive FL and data centre training. 
Surprisingly, \methodName~reaches 
an on-par privacy level as generic MPC which is (at least concerning privacy) as good as one might hope for, since for generic MPC an attack can only use publicly available information, namely the final output model. 
This means that the knowledge of several intermediate output models from different rounds does not really improve the ability of the adversary. In other words, the aggregated intermediate output models contain (already in our evaluation setup with only $15$ clients) only a small amount of individual training data.

Moreover, we observe that a larger number of rounds, and hence more leakage, results in more accurate appearance reconstructions (re-identification, visual score, and PSNR) given that the user's appearance stays the same over all training sets of a client. 
However, the later updates only improve the reconstructions slightly (if at all). 
The reason is that the gaze estimation model converges, i.e. towards the last round(s), less information could be leaked from the
target models as the effect of the training data on the model updates becomes small, and the multiple aggregation steps in FL dissipate individual information (c.f. \cref{fig:dist}).
\Cref{fig:reconstructions}-C shows the strong effect of the earlier updates introduced as auxiliary knowledge compared to a reconstruction (w/o aux.) that only relies on the last round update.

For the gaze-related properties (MAE, KL), 
\attackName~tries to reconstruct the gaze angle distribution used during training. However, in contrast to appearance, the gaze angles in the $k$-th round training set $D_{j,k}$ of a client $C_j$ do not (necessarily) depend on the previous training sets. In particular, the previously learned models (if introduced as aux.) can only slightly improve the final gaze reconstruction of $D_{j,t}$ as they
contain little information on $D_{j,t}$ due to the described convergence issue. 
Nonetheless, the auxiliary knowledge provides information about which parameters already converged.
For instance, in a late round, some training samples will no longer update the gaze estimator gradients (i.e. have a negligible effect on the training) while the previous model updates (if introduced as aux.) provide information about which parameters already converged.
\attackName~uses this knowledge by allowing weights corresponding to already converged parameters to nevertheless contribute to the reconstruction (c.f. \cref{app:convergence}).

With respect to scaling \cref{fig:reconstructions}-B shows the effect of the number of training samples per round on privacy. In general, a smaller number of training samples per round (i.e. overfitting) leaks more information with a slower convergence (i.e. more training rounds).
We also investigate the effect of our attack loss by replacing the total loss function from \cref{attack} by $L_{\text{gaze}}$ or $L_{\text{appearance}}$. For example, \cref{fig:reconstructions}-D shows how the adversary can reconstruct (transfer) the gaze of some target client (victim) to some randomly chosen face, and vice versa. Although the advantage of \methodName~over adaptive FL is larger when more private information is leaked (i.e. with our proposed total loss), such effects could be of independent interest (e.g. for image forensics such as Deepfakes \cite{westerlund2019emergence} or gaze data synthesis \cite{qin2023domainadaptive}).

\begin{table}[t]
    \centering

    \begin{tabularx}{1\textwidth} 
    { 
       >{\raggedright \arraybackslash}X 
       >{\centering\arraybackslash}X
       >{\centering\arraybackslash}X
       >{\centering\arraybackslash}X
       >{\centering\arraybackslash}X
       >{\centering\arraybackslash}X
      }
     \hline
      
       & 
       \textbf{Re-identified} &\textbf{Visual score} & \textbf{PSNR} & \textbf{MAE} & \textbf{KL} \\
     \hline
     \multirow{1}{5em}{\textbf{Data centre} } 
     &
     15/15 & 100\% &
       max. & 0 & 0\\
     \hline
     \textbf{Adaptive FL}  
     &
     15/15 & 83\% & 19.8 & 8.1 & 0.201 \\
    \hline 
    \textbf\methodName 
     & 0/15  & 7\% & 5.1 & 10.6 & 0.269 \\
     \hline
 \multirow{1}{6em}{\textbf{Generic MPC}} 
     & 
     0/15  & 3\% & 5.0 & 10.8 & 0.273 \\ 
\hline       
    \end{tabularx}
    
    \caption{\attackName~performance on the MPIIFaceGaze dataset.
    \vspace{-0.5cm}}
    \label{tab:attack}
\end{table}

In summary, since \methodName~does not leak any information on individual updates, attacks can only use the aggregated models at each round. This represents a key advantage compared to i) data centre training which requires access to the input images and ii) adaptive FL which allows access to both the individual model updates (\IU) as well as the output model (\OM) of each round. 
Results of other well-established data leakage attacks further confirm our findings (c.f. \cref{sub:attacks}).
Finally, \methodName~reaches an on-par privacy level as generic MPC, which comes with optimal privacy guarantees (under our assumption that one server is honest). However, generic MPC remains inefficient in real-world applications, e.g. gaze estimation (c.f. \cref{sub:efficency} below).

\subsection{Efficiency}\label{sub:efficency}

\begin{table}[t]
    \centering
    \begin{tabularx}{1\textwidth} 
    {  >{\centering\arraybackslash}X 
    >{\centering\arraybackslash}X 
       >{\centering\arraybackslash}X
       >{\centering\arraybackslash}X
       >{\centering\arraybackslash}X
      }
      \toprule
         & \textbf{Data Centre}  & \textbf{Adaptive FL} & \textbf{\methodName} & \textbf{Generic MPC}\\
    \midrule
    \textbf{Runtime}  & $\sim 357.0$ s & $\sim 357.2$ s & 
     $\sim 411.9$ s & $\sim 10$ months\\
     \midrule
    \textbf{Communication}  & $\sim 909$ MB & $\sim1$ MB/round & $\sim 7$ MB/round & $\sim 3840$ MB\\
    
    \bottomrule
    
    \end{tabularx}
    \caption{Training 
    runtime and communication for the baselines vs \methodName~on MPIIGaze on a NVIDIA V100 GPU. Generic MPC and \methodName~both use three servers; adaptive FL and data centre both use one server and 10 training rounds.\vspace{-0.7cm} 
   }
    \label{tab:mpc}
\end{table}
While the previous results suggest that generic MPC achieves slightly better performance and privacy than \methodName, this does not imply that generic MPC can be successfully used in practice. 
The main obstruction to employing generic MPC is that it can currently not be implemented efficiently. 
Fortunately, our combination of FL and MPC does not suffer the same disadvantage.
The reason is that the MPC computation in generic MPC is much more complex than in our FL-based scheme \methodName. In \methodName, the MPC protocol is dominated by additions of the (shares of) the individual updates. These additions can be computed locally (i.e. without communication among the servers) and therewith efficiently by each server. In contrast, a generic MPC evaluation must additionally handle all the non-linear training operations, which in \methodName~are done locally by the clients. This includes e.g. approximations of activation functions by high-degree polynomials or derivatives. By design of the MPC protocols, these non-linear operations lead to a significant communication overhead among the servers. 
For instance, 
the first convolution layer on the MPIIGaze CNN requires $\sim\!900k$ and $\sim\!4,\!000k$ communicated 128-bit values per iteration for the forward and backward pass, respectively (with a total of $\sim\!30,\!000K$ for the entire CNN). As a result, the estimated training time for three parties on a single computer 
with no network delay is $\sim\!10$ months and 3840 MB communication for training and $\sim\!4$ hrs for inference. In real-world setups, e.g. over the Internet, 
network delays and bandwidth restrictions further slow down the training (significantly). 
Overall, the generic MPC approach is currently not ready for real-world applications in gaze estimation given its runtime, as shown in \cref{tab:mpc}.
While \methodName~is much more efficient it nevertheless comes with a small efficiency loss compared to adaptive FL, since the clients need to send one share of their individual update to each server (plus some additional communication needed to get an actively secure scheme as described in \cref{sub: privatEyesTrain}). For instance \Cref{tab:mpc} shows that \methodName~for $3$ servers comes with an approx. factor $7$ communication overhead compared to adaptive FL, which is well in accordance with theory (it results from sending $3$ shares plus authentication for active security).
The communication scales (just as generic MPC) linearly in the number of servers. 
The runtime on the other hand is more dominated by the local training of the clients and not affected strongly by the communication overhead. Please note also, that shares can be sent in parallel to all servers (and back) and hence the input and output phase in adaptive FL and \methodName~have similar runtime (although more data is sent in \methodName).
We generally think that the small absolute increase in runtime and communication is justified given the significantly better privacy guarantees of \methodName~over adaptive FL.
Finally, data centre training needs significantly more communication since all the raw training data has to be sent to the central server. 

In summary, we have seen that our combination of FL and MPC in \methodName~has significant advantages compared to all other currently available gaze estimation training approaches (on distributed data sets). It comes with (i) good gaze estimation performance (which is only slightly worse than data centre training and equal to centralized FL approaches), (ii) reasonable computational and communication costs, 
(iii) strong privacy guarantees that are almost optimal and significantly better than centralised FL approaches and data centre training.

\section{Discussion}\label{sec:discussion}


In the following, we discuss the potential of \methodName along three 
axes: the privacy guarantees for the gaze data, the fairness of the gaze estimator, and the feasibility of our trust assumption. 

\noindentparagraph{Privacy guarantees for the gaze data} 
FL structurally prevents access to the client's raw data through data minimisation (i.e. the \textit{individual model updates)}. In addition, FL facilitates data-stewardship to ensure that clients control and approve of how their data will be used and have governance over their data, hence, transparency and consent principles are applied ~\cite{bonawitz2022fedprivacy}.
Unfortunately, our evaluation in \cref{sec:experiments} shows that FL is still vulnerable to attacks which can deduce very accurate reconstructions of the training data from the individual model updates. 
\methodName~does not leak the individual model updates and therefore reaches a far better privacy level.
In particular, \methodName~provides provable privacy guarantees as long as at least one server remains honest (see \cref{sec:method} for details).
Other approaches with similar privacy guarantees like generic MPC, are currently too inefficient to be applied in real-world gaze applications.

\noindentparagraph{Fairness of the gaze estimator} In collaborative approaches, e.g. FL or \methodName, concerns about model fairness are magnified due to the heterogeneous data distribution across clients. Nonetheless, our experiments (\cref{sub:results}) show that (i) \methodName~is able to adapt to the different individual model updates and yields good performance for each client (\cref{fig:individual}), (ii) increases the generalisation capability of gaze estimators, 
and (iii) maintains a high scalability performance. \methodName, therefore, incentivises collaborative training (a main goal of our paper) while preserving data privacy.

\noindentparagraph{Trust and dishonest majority} 
Trust plays a crucial role in shaping clients' 
willingness to share information, such as model updates. The evolving capabilities of AI and 
adversarial attacks, as discussed in \cref{sec:relatedwork}, have given rise to the 'trust crisis' \cite{yu17user}. Addressing this issue, \cite{steil2019privacy} conducted a comprehensive survey on users' attitudes towards sharing their eye data.
The study revealed that clients are more inclined to share their data if the co-owner, e.g. a server in PrivatEyes, belongs to entities perceived as trustworthy, such as governmental, healthcare, education, or research institutes.
Conversely, there is a lack of trust in international and profit-oriented organizations. Interestingly, clients are more open to sharing their data in aggregated forms, e.g. output models in PrivatEyes, while being reluctant to share raw data.
Hence, PrivatEyes, specifically guided by the dishonest majority assumption, emerges as a solution for (i) ensuring the trustworthy processing of eye data for any service provider (i.e. server) if only one server (i.e., the sole honest server) is affiliated with trusted entities and (ii) aligning with clients' trust dynamics.

\noindentparagraph{Limitations and future work}
One of the major challenges of collaborative gaze estimation research is to transfer conventional gaze estimation methods to decentralised datasets. 
For example, the selection of hyper-parameters and their optimisation (i.e. AutoML \cite{kohavi1995automatic}) currently does not have an efficient distributed counterpart (especially for secure setups).
Furthermore, \methodName, similar to other FL paradigms (e.g \cite{elfares2022federated}), considers a supervised gaze estimation task where data annotation is assumed to be available at each client. Applying FL (or even \methodName) to semi-supervised or unsupervised learning (i.e. \cite{yu2020unsupervised, jindal2022contrastive}) also remains an open problem. 
Finally, since federated learning raises environmental concerns and can emit more carbon than centralised training, especially with multiple servers as in PrivatEyes, we refer the reader to \cite{qiu2023first} for the choice of hardware and the corresponding energy consumption.

\section{Conclusion}
We presented \methodName{}---the first privacy-preserving training approach for appearance-based gaze estimation methods that combines FL with MPC.
Our evaluation shows that \methodName~reaches the same gaze estimation performance as the currently best (non-secure) distributed training scheme  \cite{elfares2022federated}, is domain-agnostic (i.e. can be used with any gaze estimation model and dataset), and can scale to a large number of clients,
with a rather moderate efficiency overhead compared to centralised FL. 
Finally, \methodName{}~provides strong security guarantees (cf. \cref{sec:method}).
It then reaches almost optimal privacy guarantees, if only one out of $n$ servers is honest, and is therewith significantly better than centralised FL and data centre training (cf. \cref{sec:experiments}).
Overall, \methodName~currently provides the most practical and private gaze estimation training on distributed datasets. It hits a ``sweet spot'' in terms of privacy, model accuracy, and performance. 



\mayar{Page limit is 14 pages without references.}

\begin{acks}

M. Elfares was funded by the Ministry of Science, Research and the Arts Baden-W{\"u}rttemberg in the Artificial Intelligence Software Academy (AISA).
Z. Hu was funded by the Deutsche Forschungsgemeinschaft (DFG, German Research Foundation) under Germany's Excellence Strategy - EXC 2075 – 390740016.
P. Reisert and R. K{\"u}sters were supported by the German Federal Ministry of Education and Research under Grant Agreement No. 16KIS1441 and from the French National Research Agency under Grant Agreement No. ANR-20-CYAL-0006 (CRYPTECS project).
A. Bulling was funded by the European Research Council (ERC; grant agreement 801708).

\end{acks}

\bibliographystyle{ACM-Reference-Format}
\bibliography{references}


\begin{thebibliography}{96}


\ifx \showCODEN    \undefined \def \showCODEN     #1{\unskip}     \fi
\ifx \showDOI      \undefined \def \showDOI       #1{#1}\fi
\ifx \showISBNx    \undefined \def \showISBNx     #1{\unskip}     \fi
\ifx \showISBNxiii \undefined \def \showISBNxiii  #1{\unskip}     \fi
\ifx \showISSN     \undefined \def \showISSN      #1{\unskip}     \fi
\ifx \showLCCN     \undefined \def \showLCCN      #1{\unskip}     \fi
\ifx \shownote     \undefined \def \shownote      #1{#1}          \fi
\ifx \showarticletitle \undefined \def \showarticletitle #1{#1}   \fi
\ifx \showURL      \undefined \def \showURL       {\relax}        \fi
\providecommand\bibfield[2]{#2}
\providecommand\bibinfo[2]{#2}
\providecommand\natexlab[1]{#1}
\providecommand\showeprint[2][]{arXiv:#2}

\bibitem[B{\^a}ce et~al\mbox{.}(2020)]%
        {bace20_chi}
\bibfield{author}{\bibinfo{person}{Mihai B{\^a}ce}, \bibinfo{person}{Sander
  Staal}, {and} \bibinfo{person}{Andreas Bulling}.}
  \bibinfo{year}{2020}\natexlab{}.
\newblock \showarticletitle{Quantification of Users' Visual Attention During
  Everyday Mobile Device Interactions}. In \bibinfo{booktitle}{\emph{Proc. ACM
  SIGCHI Conference on Human Factors in Computing Systems (CHI)}}.
  \bibinfo{pages}{1--14}.
\newblock
\urldef\tempurl%
\url{https://doi.org/10.1145/3313831.3376449}
\showDOI{\tempurl}


\bibitem[Baluja and Pomerleau(1993)]%
        {baluja1993non}
\bibfield{author}{\bibinfo{person}{Shumeet Baluja} {and} \bibinfo{person}{Dean
  Pomerleau}.} \bibinfo{year}{1993}\natexlab{}.
\newblock \showarticletitle{Non-intrusive gaze tracking using artificial neural
  networks}.
\newblock \bibinfo{journal}{\emph{Advances in Neural Information Processing
  Systems}}  \bibinfo{volume}{6} (\bibinfo{year}{1993}).
\newblock


\bibitem[Bernau et~al\mbox{.}(2019)]%
        {bernau2019assessing}
\bibfield{author}{\bibinfo{person}{Daniel Bernau},
  \bibinfo{person}{Philip-William Grassal}, \bibinfo{person}{Jonas Robl}, {and}
  \bibinfo{person}{Florian Kerschbaum}.} \bibinfo{year}{2019}\natexlab{}.
\newblock \showarticletitle{Assessing differentially private deep learning with
  membership inference}.
\newblock \bibinfo{journal}{\emph{arXiv preprint arXiv:1912.11328}}
  (\bibinfo{year}{2019}).
\newblock


\bibitem[Biswas et~al\mbox{.}(2021)]%
        {biswas2021appearance}
\bibfield{author}{\bibinfo{person}{Pradipta Biswas} {et~al\mbox{.}}}
  \bibinfo{year}{2021}\natexlab{}.
\newblock \showarticletitle{Appearance-based gaze estimation using attention
  and difference mechanism}. In \bibinfo{booktitle}{\emph{Proceedings of the
  IEEE/CVF conference on computer vision and pattern recognition}}.
  \bibinfo{pages}{3143--3152}.
\newblock


\bibitem[Bonawitz et~al\mbox{.}(2017)]%
        {bonawitz2017practical}
\bibfield{author}{\bibinfo{person}{Keith Bonawitz}, \bibinfo{person}{Vladimir
  Ivanov}, \bibinfo{person}{Ben Kreuter}, \bibinfo{person}{Antonio Marcedone},
  \bibinfo{person}{H~Brendan McMahan}, \bibinfo{person}{Sarvar Patel},
  \bibinfo{person}{Daniel Ramage}, \bibinfo{person}{Aaron Segal}, {and}
  \bibinfo{person}{Karn Seth}.} \bibinfo{year}{2017}\natexlab{}.
\newblock \showarticletitle{Practical secure aggregation for privacy-preserving
  machine learning}. In \bibinfo{booktitle}{\emph{proceedings of the 2017 ACM
  SIGSAC Conference on Computer and Communications Security}}.
  \bibinfo{pages}{1175--1191}.
\newblock


\bibitem[Bonawitz et~al\mbox{.}(2022)]%
        {bonawitz2022fedprivacy}
\bibfield{author}{\bibinfo{person}{Kallista Bonawitz}, \bibinfo{person}{Peter
  Kairouz}, \bibinfo{person}{Brendan Mcmahan}, {and} \bibinfo{person}{Daniel
  Ramage}.} \bibinfo{year}{2022}\natexlab{}.
\newblock \showarticletitle{Federated learning and privacy}.
\newblock \bibinfo{journal}{\emph{Commun. ACM}} \bibinfo{volume}{65},
  \bibinfo{number}{4} (\bibinfo{year}{2022}), \bibinfo{pages}{90--97}.
\newblock


\bibitem[Bozkir et~al\mbox{.}(2021)]%
        {bozkir2021diffrential}
\bibfield{author}{\bibinfo{person}{Efe Bozkir}, \bibinfo{person}{Onur
  G{\"u}nl{\"u}}, \bibinfo{person}{Wolfgang Fuhl}, \bibinfo{person}{Rafael~F
  Schaefer}, {and} \bibinfo{person}{Enkelejda Kasneci}.}
  \bibinfo{year}{2021}\natexlab{}.
\newblock \showarticletitle{Differential privacy for eye tracking with temporal
  correlations}.
\newblock \bibinfo{journal}{\emph{Plos one}} \bibinfo{volume}{16},
  \bibinfo{number}{8} (\bibinfo{year}{2021}), \bibinfo{pages}{e0255979}.
\newblock


\bibitem[Bozkir et~al\mbox{.}(2023)]%
        {bozkir2023eye}
\bibfield{author}{\bibinfo{person}{Efe Bozkir}, \bibinfo{person}{S{\"u}leyman
  {\"O}zdel}, \bibinfo{person}{Mengdi Wang}, \bibinfo{person}{Brendan
  David-John}, \bibinfo{person}{Hong Gao}, \bibinfo{person}{Kevin Butler},
  \bibinfo{person}{Eakta Jain}, {and} \bibinfo{person}{Enkelejda Kasneci}.}
  \bibinfo{year}{2023}\natexlab{}.
\newblock \showarticletitle{Eye-tracked Virtual Reality: A Comprehensive Survey
  on Methods and Privacy Challenges}.
\newblock \bibinfo{journal}{\emph{arXiv preprint arXiv:2305.14080}}
  (\bibinfo{year}{2023}).
\newblock


\bibitem[Bozkir et~al\mbox{.}(2020)]%
        {bozkir2020privacy}
\bibfield{author}{\bibinfo{person}{Efe Bozkir}, \bibinfo{person}{Ali~Burak
  {\"U}nal}, \bibinfo{person}{Mete Akg{\"u}n}, \bibinfo{person}{Enkelejda
  Kasneci}, {and} \bibinfo{person}{Nico Pfeifer}.}
  \bibinfo{year}{2020}\natexlab{}.
\newblock \showarticletitle{Privacy preserving gaze estimation using synthetic
  images via a randomized encoding based framework}. In
  \bibinfo{booktitle}{\emph{ACM symposium on eye tracking research and
  applications}}. \bibinfo{pages}{1--5}.
\newblock


\bibitem[Bulling and Gellersen(2010)]%
        {bulling2010toward}
\bibfield{author}{\bibinfo{person}{Andreas Bulling} {and} \bibinfo{person}{Hans
  Gellersen}.} \bibinfo{year}{2010}\natexlab{}.
\newblock \showarticletitle{Toward mobile eye-based human-computer
  interaction}.
\newblock \bibinfo{journal}{\emph{IEEE Pervasive Computing}}
  \bibinfo{volume}{9}, \bibinfo{number}{4} (\bibinfo{year}{2010}),
  \bibinfo{pages}{8--12}.
\newblock


\bibitem[Bulling and Roggen(2011)]%
        {bulling11_ubicomp}
\bibfield{author}{\bibinfo{person}{Andreas Bulling} {and}
  \bibinfo{person}{Daniel Roggen}.} \bibinfo{year}{2011}\natexlab{}.
\newblock \showarticletitle{Recognition of Visual Memory Recall Processes Using
  Eye Movement Analysis}. In \bibinfo{booktitle}{\emph{Proc. ACM International
  Joint Conference on Pervasive and Ubiquitous Computing (UbiComp)}}.
  \bibinfo{pages}{455--464}.
\newblock
\urldef\tempurl%
\url{https://doi.org/10.1145/2030112.2030172}
\showDOI{\tempurl}


\bibitem[Bulling et~al\mbox{.}(2013)]%
        {bulling2013eye}
\bibfield{author}{\bibinfo{person}{Andreas Bulling}, \bibinfo{person}{Christian
  Weichel}, {and} \bibinfo{person}{Hans Gellersen}.}
  \bibinfo{year}{2013}\natexlab{}.
\newblock \showarticletitle{EyeContext: Recognition of high-level contextual
  cues from human visual behaviour}. In \bibinfo{booktitle}{\emph{Proceedings
  of the sigchi conference on human factors in computing systems}}.
  \bibinfo{pages}{305--308}.
\newblock


\bibitem[Burkhalter et~al\mbox{.}(2021)]%
        {burkhalter2021rofl}
\bibfield{author}{\bibinfo{person}{Lukas Burkhalter}, \bibinfo{person}{Hidde
  Lycklama}, \bibinfo{person}{Alexander Viand}, \bibinfo{person}{Nicolas
  K{\"u}chler}, {and} \bibinfo{person}{Anwar Hithnawi}.}
  \bibinfo{year}{2021}\natexlab{}.
\newblock \showarticletitle{Rofl: Attestable robustness for secure federated
  learning}.
\newblock \bibinfo{journal}{\emph{arXiv preprint arXiv:2107.03311}}
  (\bibinfo{year}{2021}).
\newblock


\bibitem[Bâce et~al\mbox{.}(2020)]%
        {bace2020quantification}
\bibfield{author}{\bibinfo{person}{Mihai Bâce}, \bibinfo{person}{Sander
  Staal}, {and} \bibinfo{person}{Andreas Bulling}.}
  \bibinfo{year}{2020}\natexlab{}.
\newblock \showarticletitle{Quantification of Users' Visual Attention During
  Everyday Mobile Device Interactions}. In \bibinfo{booktitle}{\emph{Proc. ACM
  SIGCHI Conference on Human Factors in Computing Systems (CHI)}}.
\newblock
\urldef\tempurl%
\url{https://doi.org/10.1145/3313831.3376449}
\showDOI{\tempurl}


\bibitem[Cantoni et~al\mbox{.}(2015)]%
        {cantoni2015gant}
\bibfield{author}{\bibinfo{person}{Virginio Cantoni}, \bibinfo{person}{Chiara
  Galdi}, \bibinfo{person}{Michele Nappi}, \bibinfo{person}{Marco Porta}, {and}
  \bibinfo{person}{Daniel Riccio}.} \bibinfo{year}{2015}\natexlab{}.
\newblock \showarticletitle{GANT: Gaze analysis technique for human
  identification}.
\newblock \bibinfo{journal}{\emph{Pattern Recognition}} \bibinfo{volume}{48},
  \bibinfo{number}{4} (\bibinfo{year}{2015}), \bibinfo{pages}{1027--1038}.
\newblock


\bibitem[Chen and Ji(2008)]%
        {chen20083d}
\bibfield{author}{\bibinfo{person}{Jixu Chen} {and} \bibinfo{person}{Qiang
  Ji}.} \bibinfo{year}{2008}\natexlab{}.
\newblock \showarticletitle{3D gaze estimation with a single camera without IR
  illumination}. In \bibinfo{booktitle}{\emph{2008 19th International
  Conference on Pattern Recognition}}. IEEE, \bibinfo{pages}{1--4}.
\newblock


\bibitem[Choi et~al\mbox{.}(2013)]%
        {Choi2013appearnce}
\bibfield{author}{\bibinfo{person}{Jinsoo Choi}, \bibinfo{person}{Byungtae
  Ahn}, \bibinfo{person}{Jaesik Parl}, {and} \bibinfo{person}{In~So Kweon}.}
  \bibinfo{year}{2013}\natexlab{}.
\newblock \showarticletitle{Appearance-based gaze estimation using kinect}. In
  \bibinfo{booktitle}{\emph{2013 10th International Conference on Ubiquitous
  Robots and Ambient Intelligence}}. IEEE, \bibinfo{pages}{260--261}.
\newblock


\bibitem[Chowdhury et~al\mbox{.}(2021)]%
        {chowdhury2021eiffel}
\bibfield{author}{\bibinfo{person}{Amrita~Roy Chowdhury},
  \bibinfo{person}{Chuan Guo}, \bibinfo{person}{Somesh Jha}, {and}
  \bibinfo{person}{Laurens van~der Maaten}.} \bibinfo{year}{2021}\natexlab{}.
\newblock \showarticletitle{EIFFeL: Ensuring Integrity for Federated Learning}.
\newblock \bibinfo{journal}{\emph{arXiv preprint arXiv:2112.12727}}
  (\bibinfo{year}{2021}).
\newblock


\bibitem[Damg{\aa}rd et~al\mbox{.}(2016)]%
        {damgaard2016benchmarking}
\bibfield{author}{\bibinfo{person}{Ivan Damg{\aa}rd}, \bibinfo{person}{Kasper
  Damg{\aa}rd}, \bibinfo{person}{Kurt Nielsen},
  \bibinfo{person}{Peter~Sebastian Nordholt}, {and} \bibinfo{person}{Tomas
  Toft}.} \bibinfo{year}{2016}\natexlab{}.
\newblock \showarticletitle{Confidential benchmarking based on multiparty
  computation}. In \bibinfo{booktitle}{\emph{International Conference on
  Financial Cryptography and Data Security}}. Springer,
  \bibinfo{pages}{169--187}.
\newblock


\bibitem[Damg{\aa}rd et~al\mbox{.}(2012)]%
        {damgaard2012spdz}
\bibfield{author}{\bibinfo{person}{Ivan Damg{\aa}rd}, \bibinfo{person}{Valerio
  Pastro}, \bibinfo{person}{Nigel Smart}, {and} \bibinfo{person}{Sarah
  Zakarias}.} \bibinfo{year}{2012}\natexlab{}.
\newblock \showarticletitle{Multiparty computation from somewhat homomorphic
  encryption}. In \bibinfo{booktitle}{\emph{Annual Cryptology Conference}}.
  Springer, \bibinfo{pages}{643--662}.
\newblock


\bibitem[Dong et~al\mbox{.}(2021)]%
        {dong2021flod}
\bibfield{author}{\bibinfo{person}{Ye Dong}, \bibinfo{person}{Xiaojun Chen},
  \bibinfo{person}{Kaiyun Li}, \bibinfo{person}{Dakui Wang}, {and}
  \bibinfo{person}{Shuai Zeng}.} \bibinfo{year}{2021}\natexlab{}.
\newblock \showarticletitle{FLOD: Oblivious defender for private
  Byzantine-robust federated learning with dishonest-majority}. In
  \bibinfo{booktitle}{\emph{European Symposium on Research in Computer
  Security}}. Springer, \bibinfo{pages}{497--518}.
\newblock


\bibitem[Dwork et~al\mbox{.}(2014)]%
        {dwork2014algorithmic}
\bibfield{author}{\bibinfo{person}{Cynthia Dwork}, \bibinfo{person}{Aaron
  Roth}, {et~al\mbox{.}}} \bibinfo{year}{2014}\natexlab{}.
\newblock \showarticletitle{The algorithmic foundations of differential
  privacy}.
\newblock \bibinfo{journal}{\emph{Foundations and Trends{\textregistered} in
  Theoretical Computer Science}} \bibinfo{volume}{9}, \bibinfo{number}{3--4}
  (\bibinfo{year}{2014}), \bibinfo{pages}{211--407}.
\newblock


\bibitem[Elfares et~al\mbox{.}(2022)]%
        {elfares2022federated}
\bibfield{author}{\bibinfo{person}{Mayar Elfares}, \bibinfo{person}{Zhiming
  Hu}, \bibinfo{person}{Pascal Reisert}, \bibinfo{person}{Andreas Bulling},
  {and} \bibinfo{person}{Ralf K{\"u}sters}.} \bibinfo{year}{2022}\natexlab{}.
\newblock \showarticletitle{Federated Learning for Appearance-based Gaze
  Estimation in the Wild}. In \bibinfo{booktitle}{\emph{Proc. NeurIPS Workshop
  on Gaze Meets ML (GMML)}}. \bibinfo{pages}{1--11}.
\newblock


\bibitem[Faber et~al\mbox{.}(2018)]%
        {faber2018automated}
\bibfield{author}{\bibinfo{person}{Myrthe Faber}, \bibinfo{person}{Robert
  Bixler}, {and} \bibinfo{person}{Sidney~K D’Mello}.}
  \bibinfo{year}{2018}\natexlab{}.
\newblock \showarticletitle{An automated behavioral measure of mind wandering
  during computerized reading}.
\newblock \bibinfo{journal}{\emph{Behavior Research Methods}}
  \bibinfo{volume}{50}, \bibinfo{number}{1} (\bibinfo{year}{2018}),
  \bibinfo{pages}{134--150}.
\newblock


\bibitem[Fereidooni et~al\mbox{.}(2021)]%
        {fereidooni2021safelearn}
\bibfield{author}{\bibinfo{person}{Hossein Fereidooni}, \bibinfo{person}{Samuel
  Marchal}, \bibinfo{person}{Markus Miettinen}, \bibinfo{person}{Azalia
  Mirhoseini}, \bibinfo{person}{Helen M{\"o}llering},
  \bibinfo{person}{Thien~Duc Nguyen}, \bibinfo{person}{Phillip Rieger},
  \bibinfo{person}{Ahmad-Reza Sadeghi}, \bibinfo{person}{Thomas Schneider},
  \bibinfo{person}{Hossein Yalame}, {et~al\mbox{.}}}
  \bibinfo{year}{2021}\natexlab{}.
\newblock \showarticletitle{SAFELearn: Secure aggregation for private federated
  learning}. In \bibinfo{booktitle}{\emph{2021 IEEE Security and Privacy
  Workshops (SPW)}}. IEEE, \bibinfo{pages}{56--62}.
\newblock


\bibitem[Fredrikson et~al\mbox{.}(2015)]%
        {fredrikson2015model}
\bibfield{author}{\bibinfo{person}{Matt Fredrikson}, \bibinfo{person}{Somesh
  Jha}, {and} \bibinfo{person}{Thomas Ristenpart}.}
  \bibinfo{year}{2015}\natexlab{}.
\newblock \showarticletitle{Model inversion attacks that exploit confidence
  information and basic countermeasures}. In \bibinfo{booktitle}{\emph{ACM
  CCS}}. \bibinfo{pages}{1322--1333}.
\newblock


\bibitem[Galli et~al\mbox{.}(2022)]%
        {galli2022machine}
\bibfield{author}{\bibinfo{person}{Filippo Galli}, \bibinfo{person}{Sayan
  Biswas}, \bibinfo{person}{Kangsoo Jung}, \bibinfo{person}{Tommaso Cucinotta},
  {and} \bibinfo{person}{Catuscia Palamidessi}.}
  \bibinfo{year}{2022}\natexlab{}.
\newblock \bibinfo{title}{Group privacy for personalized federated learning}.
\newblock
\newblock
\urldef\tempurl%
\url{https://doi.org/10.48550/ARXIV.2206.03396}
\showDOI{\tempurl}


\bibitem[Geiping et~al\mbox{.}(2020)]%
        {geiping2020inverting}
\bibfield{author}{\bibinfo{person}{Jonas Geiping}, \bibinfo{person}{Hartmut
  Bauermeister}, \bibinfo{person}{Hannah Dr{\"o}ge}, {and}
  \bibinfo{person}{Michael Moeller}.} \bibinfo{year}{2020}\natexlab{}.
\newblock \showarticletitle{Inverting gradients-how easy is it to break privacy
  in federated learning?}
\newblock \bibinfo{journal}{\emph{Advances in Neural Information Processing
  Systems}}  \bibinfo{volume}{33} (\bibinfo{year}{2020}),
  \bibinfo{pages}{16937--16947}.
\newblock


\bibitem[Gilad-Bachrach et~al\mbox{.}(2016)]%
        {gilad2016cryptonets}
\bibfield{author}{\bibinfo{person}{Ran Gilad-Bachrach}, \bibinfo{person}{Nathan
  Dowlin}, \bibinfo{person}{Kim Laine}, \bibinfo{person}{Kristin Lauter},
  \bibinfo{person}{Michael Naehrig}, {and} \bibinfo{person}{John Wernsing}.}
  \bibinfo{year}{2016}\natexlab{}.
\newblock \showarticletitle{Cryptonets: Applying neural networks to encrypted
  data with high throughput and accuracy}. In
  \bibinfo{booktitle}{\emph{International conference on machine learning}}.
  PMLR, \bibinfo{pages}{201--210}.
\newblock


\bibitem[Hansen and Ji(2009)]%
        {hansen2009beholder}
\bibfield{author}{\bibinfo{person}{Dan~Witzner Hansen} {and}
  \bibinfo{person}{Qiang Ji}.} \bibinfo{year}{2009}\natexlab{}.
\newblock \showarticletitle{In the eye of the beholder: A survey of models for
  eyes and gaze}.
\newblock \bibinfo{journal}{\emph{IEEE transactions on pattern analysis and
  machine intelligence}} \bibinfo{volume}{32}, \bibinfo{number}{3}
  (\bibinfo{year}{2009}), \bibinfo{pages}{478--500}.
\newblock


\bibitem[He et~al\mbox{.}(2019)]%
        {he2019model}
\bibfield{author}{\bibinfo{person}{Zecheng He}, \bibinfo{person}{Tianwei
  Zhang}, {and} \bibinfo{person}{Ruby~B Lee}.} \bibinfo{year}{2019}\natexlab{}.
\newblock \showarticletitle{Model inversion attacks against collaborative
  inference}. In \bibinfo{booktitle}{\emph{Proceedings of the 35th Annual
  Computer Security Applications Conference}}. \bibinfo{pages}{148--162}.
\newblock


\bibitem[Hennessey et~al\mbox{.}(2006)]%
        {hennessey2006single}
\bibfield{author}{\bibinfo{person}{Craig Hennessey}, \bibinfo{person}{Borna
  Noureddin}, {and} \bibinfo{person}{Peter Lawrence}.}
  \bibinfo{year}{2006}\natexlab{}.
\newblock \showarticletitle{A single camera eye-gaze tracking system with free
  head motion}. In \bibinfo{booktitle}{\emph{Proceedings of the 2006 symposium
  on Eye tracking research \& applications}}. \bibinfo{pages}{87--94}.
\newblock


\bibitem[Higuchi et~al\mbox{.}(2018)]%
        {higuchi2018visualizing}
\bibfield{author}{\bibinfo{person}{Keita Higuchi}, \bibinfo{person}{Soichiro
  Matsuda}, \bibinfo{person}{Rie Kamikubo}, \bibinfo{person}{Takuya Enomoto},
  \bibinfo{person}{Yusuke Sugano}, \bibinfo{person}{Junichi Yamamoto}, {and}
  \bibinfo{person}{Yoichi Sato}.} \bibinfo{year}{2018}\natexlab{}.
\newblock \showarticletitle{Visualizing gaze direction to support video coding
  of social attention for children with autism spectrum disorder}. In
  \bibinfo{booktitle}{\emph{23rd International Conference on Intelligent User
  Interfaces}}. \bibinfo{pages}{571--582}.
\newblock


\bibitem[Hitaj et~al\mbox{.}(2017)]%
        {hitaj2017deep}
\bibfield{author}{\bibinfo{person}{Briland Hitaj}, \bibinfo{person}{Giuseppe
  Ateniese}, {and} \bibinfo{person}{Fernando Perez-Cruz}.}
  \bibinfo{year}{2017}\natexlab{}.
\newblock \showarticletitle{Deep models under the GAN: information leakage from
  collaborative deep learning}. In \bibinfo{booktitle}{\emph{Proceedings of the
  2017 ACM SIGSAC conference on computer and communications security}}.
  \bibinfo{pages}{603--618}.
\newblock


\bibitem[Holzman et~al\mbox{.}(1974)]%
        {holzman1974eye}
\bibfield{author}{\bibinfo{person}{Philip~S Holzman},
  \bibinfo{person}{Leonard~R Proctor}, \bibinfo{person}{Deborah~L Levy},
  \bibinfo{person}{Nicholas~J Yasillo}, \bibinfo{person}{Herbert~Y Meltzer},
  {and} \bibinfo{person}{Stephen~W Hurt}.} \bibinfo{year}{1974}\natexlab{}.
\newblock \showarticletitle{Eye-tracking dysfunctions in schizophrenic patients
  and their relatives}.
\newblock \bibinfo{journal}{\emph{Archives of general psychiatry}}
  \bibinfo{volume}{31}, \bibinfo{number}{2} (\bibinfo{year}{1974}),
  \bibinfo{pages}{143--151}.
\newblock


\bibitem[Hoppe et~al\mbox{.}(2018)]%
        {hoppe2018eye}
\bibfield{author}{\bibinfo{person}{Sabrina Hoppe}, \bibinfo{person}{Tobias
  Loetscher}, \bibinfo{person}{Stephanie~A Morey}, {and}
  \bibinfo{person}{Andreas Bulling}.} \bibinfo{year}{2018}\natexlab{}.
\newblock \showarticletitle{Eye movements during everyday behavior predict
  personality traits}.
\newblock \bibinfo{journal}{\emph{Frontiers in Human Neuroscience}}
  (\bibinfo{year}{2018}), \bibinfo{pages}{105}.
\newblock


\bibitem[Huang et~al\mbox{.}(2016a)]%
        {huang2016building}
\bibfield{author}{\bibinfo{person}{Michael~Xuelin Huang},
  \bibinfo{person}{Tiffany~CK Kwok}, \bibinfo{person}{Grace Ngai},
  \bibinfo{person}{Stephen~CF Chan}, {and} \bibinfo{person}{Hong~Va Leong}.}
  \bibinfo{year}{2016}\natexlab{a}.
\newblock \showarticletitle{Building a personalized, auto-calibrating eye
  tracker from user interactions}. In \bibinfo{booktitle}{\emph{Proceedings of
  the 2016 CHI Conference on Human Factors in Computing Systems}}.
  \bibinfo{pages}{5169--5179}.
\newblock


\bibitem[Huang et~al\mbox{.}(2016b)]%
        {huang2016stress}
\bibfield{author}{\bibinfo{person}{Michael~Xuelin Huang},
  \bibinfo{person}{Jiajia Li}, \bibinfo{person}{Grace Ngai}, {and}
  \bibinfo{person}{Hong~Va Leong}.} \bibinfo{year}{2016}\natexlab{b}.
\newblock \showarticletitle{Stressclick: Sensing stress from gaze-click
  patterns}. In \bibinfo{booktitle}{\emph{Proceedings of the 24th ACM
  international conference on Multimedia}}. \bibinfo{pages}{1395--1404}.
\newblock


\bibitem[Huang et~al\mbox{.}(2017)]%
        {huang2017screen}
\bibfield{author}{\bibinfo{person}{Michael~Xuelin Huang},
  \bibinfo{person}{Jiajia Li}, \bibinfo{person}{Grace Ngai}, {and}
  \bibinfo{person}{Hong~Va Leong}.} \bibinfo{year}{2017}\natexlab{}.
\newblock \showarticletitle{Screenglint: Practical, in-situ gaze estimation on
  smartphones}. In \bibinfo{booktitle}{\emph{Proceedings of the 2017 CHI
  Conference on Human Factors in Computing Systems}}.
  \bibinfo{pages}{2546--2557}.
\newblock


\bibitem[Hutton et~al\mbox{.}(1984)]%
        {hutton1984eye}
\bibfield{author}{\bibinfo{person}{J~Thomas Hutton}, \bibinfo{person}{JA
  Nagel}, {and} \bibinfo{person}{Ruth~B Loewenson}.}
  \bibinfo{year}{1984}\natexlab{}.
\newblock \showarticletitle{Eye tracking dysfunction in Alzheimer-type
  dementia}.
\newblock \bibinfo{journal}{\emph{Neurology}} \bibinfo{volume}{34},
  \bibinfo{number}{1} (\bibinfo{year}{1984}), \bibinfo{pages}{99--99}.
\newblock


\bibitem[Jindal and Manduchi(2022)]%
        {jindal2022contrastive}
\bibfield{author}{\bibinfo{person}{Swati Jindal} {and} \bibinfo{person}{Roberto
  Manduchi}.} \bibinfo{year}{2022}\natexlab{}.
\newblock \bibinfo{title}{Contrastive Representation Learning for Gaze
  Estimation}.
\newblock
\newblock


\bibitem[Kairouz et~al\mbox{.}(2019)]%
        {kairouz2019advances}
\bibfield{author}{\bibinfo{person}{Peter Kairouz}, \bibinfo{person}{H.~Brendan
  McMahan}, \bibinfo{person}{Brendan Avent}, \bibinfo{person}{Aurélien
  Bellet}, \bibinfo{person}{Mehdi Bennis}, \bibinfo{person}{Arjun~Nitin
  Bhagoji}, \bibinfo{person}{K.~A. Bonawitz}, \bibinfo{person}{Zachary
  Charles}, \bibinfo{person}{Graham Cormode}, \bibinfo{person}{Rachel
  Cummings}, \bibinfo{person}{Rafael~G.L. D'Oliveira},
  \bibinfo{person}{Salim~El Rouayheb}, \bibinfo{person}{David Evans},
  \bibinfo{person}{Josh Gardner}, \bibinfo{person}{Zachary Garrett},
  \bibinfo{person}{Adrià Gascón}, \bibinfo{person}{Badih Ghazi},
  \bibinfo{person}{Phillip~B. Gibbons}, \bibinfo{person}{Marco Gruteser},
  \bibinfo{person}{Zaid Harchaoui}, \bibinfo{person}{Chaoyang He},
  \bibinfo{person}{Lie He}, \bibinfo{person}{Zhouyuan Huo},
  \bibinfo{person}{Ben Hutchinson}, \bibinfo{person}{Justin Hsu},
  \bibinfo{person}{Martin Jaggi}, \bibinfo{person}{Tara Javidi},
  \bibinfo{person}{Gauri Joshi}, \bibinfo{person}{Mikhail Khodak},
  \bibinfo{person}{Jakub Konečný}, \bibinfo{person}{Aleksandra Korolova},
  \bibinfo{person}{Farinaz Koushanfar}, \bibinfo{person}{Sanmi Koyejo},
  \bibinfo{person}{Tancrède Lepoint}, \bibinfo{person}{Yang Liu},
  \bibinfo{person}{Prateek Mittal}, \bibinfo{person}{Mehryar Mohri},
  \bibinfo{person}{Richard Nock}, \bibinfo{person}{Ayfer Özgür},
  \bibinfo{person}{Rasmus Pagh}, \bibinfo{person}{Mariana Raykova},
  \bibinfo{person}{Hang Qi}, \bibinfo{person}{Daniel Ramage},
  \bibinfo{person}{Ramesh Raskar}, \bibinfo{person}{Dawn Song},
  \bibinfo{person}{Weikang Song}, \bibinfo{person}{Sebastian~U. Stich},
  \bibinfo{person}{Ziteng Sun}, \bibinfo{person}{Ananda~Theertha Suresh},
  \bibinfo{person}{Florian Tramèr}, \bibinfo{person}{Praneeth Vepakomma},
  \bibinfo{person}{Jianyu Wang}, \bibinfo{person}{Li Xiong},
  \bibinfo{person}{Zheng Xu}, \bibinfo{person}{Qiang Yang},
  \bibinfo{person}{Felix~X. Yu}, \bibinfo{person}{Han Yu}, {and}
  \bibinfo{person}{Sen Zhao}.} \bibinfo{year}{2019}\natexlab{}.
\newblock \showarticletitle{Advances and Open Problems in Federated Learning}.
\newblock  (\bibinfo{year}{2019}).
\newblock
\urldef\tempurl%
\url{https://arxiv.org/abs/1912.04977}
\showURL{%
\tempurl}


\bibitem[Keller(2020)]%
        {keller2020mpspdz}
\bibfield{author}{\bibinfo{person}{Marcel Keller}.}
  \bibinfo{year}{2020}\natexlab{}.
\newblock \showarticletitle{MP-SPDZ: A versatile framework for multi-party
  computation}. In \bibinfo{booktitle}{\emph{Proceedings of the 2020 ACM SIGSAC
  conference on computer and communications security}}.
  \bibinfo{pages}{1575--1590}.
\newblock


\bibitem[Keller et~al\mbox{.}(2018)]%
        {keller2018overdrive}
\bibfield{author}{\bibinfo{person}{Marcel Keller}, \bibinfo{person}{Valerio
  Pastro}, {and} \bibinfo{person}{Dragos Rotaru}.}
  \bibinfo{year}{2018}\natexlab{}.
\newblock \showarticletitle{Overdrive: making SPDZ great again}. In
  \bibinfo{booktitle}{\emph{Annual International Conference on the Theory and
  Applications of Cryptographic Techniques}}. Springer,
  \bibinfo{pages}{158--189}.
\newblock


\bibitem[Kifer and Machanavajjhala(2011)]%
        {kifer2011no}
\bibfield{author}{\bibinfo{person}{Daniel Kifer} {and} \bibinfo{person}{Ashwin
  Machanavajjhala}.} \bibinfo{year}{2011}\natexlab{}.
\newblock \showarticletitle{No free lunch in data privacy}. In
  \bibinfo{booktitle}{\emph{Proceedings of the 2011 ACM SIGMOD International
  Conference on Management of data}}. \bibinfo{pages}{193--204}.
\newblock


\bibitem[Kim et~al\mbox{.}(2019)]%
        {kim2019nvgaze}
\bibfield{author}{\bibinfo{person}{Joohwan Kim}, \bibinfo{person}{Michael
  Stengel}, \bibinfo{person}{Alexander Majercik}, \bibinfo{person}{Shalini
  De~Mello}, \bibinfo{person}{David Dunn}, \bibinfo{person}{Samuli Laine},
  \bibinfo{person}{Morgan McGuire}, {and} \bibinfo{person}{David Luebke}.}
  \bibinfo{year}{2019}\natexlab{}.
\newblock \showarticletitle{Nvgaze: An anatomically-informed dataset for
  low-latency, near-eye gaze estimation}. In
  \bibinfo{booktitle}{\emph{Proceedings of the 2019 CHI conference on human
  factors in computing systems}}. \bibinfo{pages}{1--12}.
\newblock


\bibitem[Kohavi and John(1995)]%
        {kohavi1995automatic}
\bibfield{author}{\bibinfo{person}{Ron Kohavi} {and} \bibinfo{person}{George~H
  John}.} \bibinfo{year}{1995}\natexlab{}.
\newblock \showarticletitle{Automatic parameter selection by minimizing
  estimated error}.
\newblock In \bibinfo{booktitle}{\emph{Machine Learning Proceedings 1995}}.
  \bibinfo{publisher}{Elsevier}, \bibinfo{pages}{304--312}.
\newblock


\bibitem[Krafka et~al\mbox{.}(2016)]%
        {krafka2016eye}
\bibfield{author}{\bibinfo{person}{Kyle Krafka}, \bibinfo{person}{Aditya
  Khosla}, \bibinfo{person}{Petr Kellnhofer}, \bibinfo{person}{Harini Kannan},
  \bibinfo{person}{Suchendra Bhandarkar}, \bibinfo{person}{Wojciech Matusik},
  {and} \bibinfo{person}{Antonio Torralba}.} \bibinfo{year}{2016}\natexlab{}.
\newblock \showarticletitle{Eye tracking for everyone}. In
  \bibinfo{booktitle}{\emph{Proceedings of the IEEE conference on computer
  vision and pattern recognition}}. \bibinfo{pages}{2176--2184}.
\newblock


\bibitem[Li et~al\mbox{.}(2022)]%
        {li2022appearance}
\bibfield{author}{\bibinfo{person}{Jing Li}, \bibinfo{person}{Zejin Chen},
  \bibinfo{person}{Yihao Zhong}, \bibinfo{person}{Hak-Keung Lam},
  \bibinfo{person}{Junxia Han}, \bibinfo{person}{Gaoxiang Ouyang},
  \bibinfo{person}{Xiaoli Li}, {and} \bibinfo{person}{Honghai Liu}.}
  \bibinfo{year}{2022}\natexlab{}.
\newblock \showarticletitle{Appearance-Based Gaze Estimation for ASD
  Diagnosis}.
\newblock \bibinfo{journal}{\emph{IEEE Transactions on Cybernetics}}
  (\bibinfo{year}{2022}).
\newblock


\bibitem[Li et~al\mbox{.}(2021)]%
        {li2021kaleido}
\bibfield{author}{\bibinfo{person}{Jingjie Li}, \bibinfo{person}{Amrita~Roy
  Chowdhury}, \bibinfo{person}{Kassem Fawaz}, {and} \bibinfo{person}{Younghyun
  Kim}.} \bibinfo{year}{2021}\natexlab{}.
\newblock \showarticletitle{$\{$Kal$\varepsilon$ido$\}$:$\{$Real-Time$\}$
  Privacy Control for $\{$Eye-Tracking$\}$ Systems}. In
  \bibinfo{booktitle}{\emph{30th USENIX Security Symposium}}.
  \bibinfo{pages}{1793--1810}.
\newblock


\bibitem[Li et~al\mbox{.}(2020)]%
        {li2020membership}
\bibfield{author}{\bibinfo{person}{Jiacheng Li}, \bibinfo{person}{Ninghui Li},
  {and} \bibinfo{person}{Bruno Ribeiro}.} \bibinfo{year}{2020}\natexlab{}.
\newblock \showarticletitle{Membership inference attacks and defenses in
  supervised learning via generalization gap}.
\newblock \bibinfo{journal}{\emph{ArXiv}}  \bibinfo{volume}{abs/2002.12062}
  (\bibinfo{year}{2020}).
\newblock


\bibitem[Liang et~al\mbox{.}(2013)]%
        {liang2013appearnce}
\bibfield{author}{\bibinfo{person}{Ke Liang}, \bibinfo{person}{Youssef Chahir},
  \bibinfo{person}{Michele Molina}, \bibinfo{person}{Charles Tijus}, {and}
  \bibinfo{person}{Fran{\c{c}}ois Jouen}.} \bibinfo{year}{2013}\natexlab{}.
\newblock \showarticletitle{Appearance-based gaze tracking with spectral
  clustering and semi-supervised gaussian process regression}. In
  \bibinfo{booktitle}{\emph{Proceedings of the 2013 Conference on Eye Tracking
  South Africa}}. \bibinfo{pages}{17--23}.
\newblock


\bibitem[Liu et~al\mbox{.}(2019)]%
        {liu2019differential}
\bibfield{author}{\bibinfo{person}{Ao Liu}, \bibinfo{person}{Lirong Xia},
  \bibinfo{person}{Andrew Duchowski}, \bibinfo{person}{Reynold Bailey},
  \bibinfo{person}{Kenneth Holmqvist}, {and} \bibinfo{person}{Eakta Jain}.}
  \bibinfo{year}{2019}\natexlab{}.
\newblock \showarticletitle{Differential privacy for eye-tracking data}. In
  \bibinfo{booktitle}{\emph{ACM Symposium on Eye Tracking Research \&
  Applications}}. \bibinfo{pages}{1--10}.
\newblock


\bibitem[Liu et~al\mbox{.}(2017)]%
        {liu2017minionn}
\bibfield{author}{\bibinfo{person}{Jian Liu}, \bibinfo{person}{Mika Juuti},
  \bibinfo{person}{Yao Lu}, {and} \bibinfo{person}{Nadarajah Asokan}.}
  \bibinfo{year}{2017}\natexlab{}.
\newblock \showarticletitle{Oblivious neural network predictions via minionn
  transformations}. In \bibinfo{booktitle}{\emph{Proceedings of the 2017 ACM
  SIGSAC conference on computer and communications security}}.
  \bibinfo{pages}{619--631}.
\newblock


\bibitem[Liu et~al\mbox{.}(2022b)]%
        {liu2022threats}
\bibfield{author}{\bibinfo{person}{Pengrui Liu}, \bibinfo{person}{Xiangrui Xu},
  {and} \bibinfo{person}{Wei Wang}.} \bibinfo{year}{2022}\natexlab{b}.
\newblock \showarticletitle{Threats, attacks and defenses to federated
  learning: issues, taxonomy and perspectives}.
\newblock \bibinfo{journal}{\emph{Cybersecurity}} \bibinfo{volume}{5},
  \bibinfo{number}{1} (\bibinfo{year}{2022}), \bibinfo{pages}{1--19}.
\newblock


\bibitem[Liu et~al\mbox{.}(2022a)]%
        {liu2022mldoc}
\bibfield{author}{\bibinfo{person}{Yugeng Liu}, \bibinfo{person}{Rui Wen},
  \bibinfo{person}{Xinlei He}, \bibinfo{person}{Ahmed Salem},
  \bibinfo{person}{Zhikun Zhang}, \bibinfo{person}{Michael Backes},
  \bibinfo{person}{Emiliano De~Cristofaro}, \bibinfo{person}{Mario Fritz},
  {and} \bibinfo{person}{Yang Zhang}.} \bibinfo{year}{2022}\natexlab{a}.
\newblock \showarticletitle{$\{$ML-Doctor$\}$: Holistic Risk Assessment of
  Inference Attacks Against Machine Learning Models}. In
  \bibinfo{booktitle}{\emph{31st USENIX Security Symposium (USENIX Security
  22)}}. \bibinfo{pages}{4525--4542}.
\newblock


\bibitem[McMahan et~al\mbox{.}(2017)]%
        {mcmahan2017fl}
\bibfield{author}{\bibinfo{person}{Brendan McMahan}, \bibinfo{person}{Eider
  Moore}, \bibinfo{person}{Daniel Ramage}, \bibinfo{person}{Seth Hampson},
  {and} \bibinfo{person}{Blaise~Aguera y Arcas}.}
  \bibinfo{year}{2017}\natexlab{}.
\newblock \showarticletitle{Communication-efficient learning of deep networks
  from decentralized data}. In \bibinfo{booktitle}{\emph{Artificial
  intelligence and statistics}}. PMLR, \bibinfo{pages}{1273--1282}.
\newblock


\bibitem[Melis et~al\mbox{.}(2019)]%
        {melis2019exploiting}
\bibfield{author}{\bibinfo{person}{Luca Melis}, \bibinfo{person}{Congzheng
  Song}, \bibinfo{person}{Emiliano De~Cristofaro}, {and}
  \bibinfo{person}{Vitaly Shmatikov}.} \bibinfo{year}{2019}\natexlab{}.
\newblock \showarticletitle{Exploiting unintended feature leakage in
  collaborative learning}. In \bibinfo{booktitle}{\emph{2019 IEEE symposium on
  security and privacy (SP)}}. IEEE, \bibinfo{pages}{691--706}.
\newblock


\bibitem[Mohassel and Rindal(2018)]%
        {mohassel2018aby3}
\bibfield{author}{\bibinfo{person}{Payman Mohassel} {and}
  \bibinfo{person}{Peter Rindal}.} \bibinfo{year}{2018}\natexlab{}.
\newblock \showarticletitle{ABY3: A mixed protocol framework for machine
  learning}. In \bibinfo{booktitle}{\emph{Proceedings of the 2018 ACM SIGSAC
  conference on computer and communications security}}.
  \bibinfo{pages}{35--52}.
\newblock


\bibitem[Mohassel and Zhang(2017)]%
        {mohassel2017secureml}
\bibfield{author}{\bibinfo{person}{Payman Mohassel} {and}
  \bibinfo{person}{Yupeng Zhang}.} \bibinfo{year}{2017}\natexlab{}.
\newblock \showarticletitle{Secureml: A system for scalable privacy-preserving
  machine learning}. In \bibinfo{booktitle}{\emph{2017 IEEE symposium on
  security and privacy (SP)}}. IEEE, \bibinfo{pages}{19--38}.
\newblock


\bibitem[Morimoto et~al\mbox{.}(2002)]%
        {morimoto2002detecting}
\bibfield{author}{\bibinfo{person}{Carlos~Hitoshi Morimoto},
  \bibinfo{person}{Arnon Amir}, {and} \bibinfo{person}{Myron Flickner}.}
  \bibinfo{year}{2002}\natexlab{}.
\newblock \showarticletitle{Detecting eye position and gaze from a single
  camera and 2 light sources}. In \bibinfo{booktitle}{\emph{2002 International
  Conference on Pattern Recognition}}, Vol.~\bibinfo{volume}{4}. IEEE,
  \bibinfo{pages}{314--317}.
\newblock


\bibitem[Nguyen et~al\mbox{.}(2022)]%
        {nguyen2022flame}
\bibfield{author}{\bibinfo{person}{Thien~Duc Nguyen}, \bibinfo{person}{Phillip
  Rieger}, \bibinfo{person}{Roberta De~Viti}, \bibinfo{person}{Huili Chen},
  \bibinfo{person}{Bj{\"o}rn~B Brandenburg}, \bibinfo{person}{Hossein Yalame},
  \bibinfo{person}{Helen M{\"o}llering}, \bibinfo{person}{Hossein Fereidooni},
  \bibinfo{person}{Samuel Marchal}, \bibinfo{person}{Markus Miettinen},
  {et~al\mbox{.}}} \bibinfo{year}{2022}\natexlab{}.
\newblock \showarticletitle{$\{$FLAME$\}$: Taming backdoors in federated
  learning}. In \bibinfo{booktitle}{\emph{31st USENIX Security Symposium
  (USENIX Security 22)}}. \bibinfo{pages}{1415--1432}.
\newblock


\bibitem[Nissim et~al\mbox{.}(2007)]%
        {nissim2007smooth}
\bibfield{author}{\bibinfo{person}{Kobbi Nissim}, \bibinfo{person}{Sofya
  Raskhodnikova}, {and} \bibinfo{person}{Adam Smith}.}
  \bibinfo{year}{2007}\natexlab{}.
\newblock \showarticletitle{Smooth sensitivity and sampling in private data
  analysis}. In \bibinfo{booktitle}{\emph{ACM Symposium on Theory of
  Computing}}. \bibinfo{pages}{75--84}.
\newblock


\bibitem[Paletta et~al\mbox{.}(2020)]%
        {paletta2020mira}
\bibfield{author}{\bibinfo{person}{Lucas Paletta}, \bibinfo{person}{Martin
  Pszeida}, \bibinfo{person}{Amir Dini}, \bibinfo{person}{Silvia Russegger},
  \bibinfo{person}{Sandra Schuessler}, \bibinfo{person}{Anna Jos},
  \bibinfo{person}{Eva Schuster}, \bibinfo{person}{Josef Steiner}, {and}
  \bibinfo{person}{Maria Fellner}.} \bibinfo{year}{2020}\natexlab{}.
\newblock \showarticletitle{MIRA--A Gaze-based Serious Game for Continuous
  Estimation of Alzheimer's Mental State}. In \bibinfo{booktitle}{\emph{ACM
  Symposium on Eye Tracking Research and Applications}}. \bibinfo{pages}{1--3}.
\newblock


\bibitem[Qin et~al\mbox{.}(2023)]%
        {qin2023domainadaptive}
\bibfield{author}{\bibinfo{person}{Jiawei Qin}, \bibinfo{person}{Takuru
  Shimoyama}, \bibinfo{person}{Xucong Zhang}, {and} \bibinfo{person}{Yusuke
  Sugano}.} \bibinfo{year}{2023}\natexlab{}.
\newblock \bibinfo{title}{Domain-Adaptive Full-Face Gaze Estimation via
  Novel-View-Synthesis and Feature Disentanglement}.
\newblock
\newblock
\showeprint[arxiv]{2305.16140}~[cs.CV]


\bibitem[Qiu et~al\mbox{.}(2023)]%
        {qiu2023first}
\bibfield{author}{\bibinfo{person}{Xinchi Qiu}, \bibinfo{person}{Titouan
  Parcollet}, \bibinfo{person}{Javier Fernandez-Marques},
  \bibinfo{person}{Pedro~PB Gusmao}, \bibinfo{person}{Yan Gao},
  \bibinfo{person}{Daniel~J Beutel}, \bibinfo{person}{Taner Topal},
  \bibinfo{person}{Akhil Mathur}, {and} \bibinfo{person}{Nicholas~D Lane}.}
  \bibinfo{year}{2023}\natexlab{}.
\newblock \showarticletitle{A first look into the carbon footprint of federated
  learning}.
\newblock \bibinfo{journal}{\emph{Journal of Machine Learning Research}}
  \bibinfo{volume}{24}, \bibinfo{number}{129} (\bibinfo{year}{2023}),
  \bibinfo{pages}{1--23}.
\newblock


\bibitem[Rathee et~al\mbox{.}(2023)]%
        {rathee2023elsa}
\bibfield{author}{\bibinfo{person}{Mayank Rathee}, \bibinfo{person}{Conghao
  Shen}, \bibinfo{person}{Sameer Wagh}, {and} \bibinfo{person}{Raluca~Ada
  Popa}.} \bibinfo{year}{2023}\natexlab{}.
\newblock \showarticletitle{Elsa: Secure aggregation for federated learning
  with malicious actors}. In \bibinfo{booktitle}{\emph{2023 IEEE Symposium on
  Security and Privacy (SP)}}. IEEE, \bibinfo{pages}{1961--1979}.
\newblock


\bibitem[Reddi et~al\mbox{.}(2020)]%
        {reddi2020adaptive}
\bibfield{author}{\bibinfo{person}{Sashank Reddi}, \bibinfo{person}{Zachary
  Charles}, \bibinfo{person}{Manzil Zaheer}, \bibinfo{person}{Zachary Garrett},
  \bibinfo{person}{Keith Rush}, \bibinfo{person}{Jakub Kone{\v{c}}n{\`y}},
  \bibinfo{person}{Sanjiv Kumar}, {and} \bibinfo{person}{H~Brendan McMahan}.}
  \bibinfo{year}{2020}\natexlab{}.
\newblock \showarticletitle{Adaptive federated optimization}.
\newblock \bibinfo{journal}{\emph{arXiv preprint arXiv:2003.00295}}
  (\bibinfo{year}{2020}).
\newblock


\bibitem[Rouhani et~al\mbox{.}(2018)]%
        {rouhani2018deepsecure}
\bibfield{author}{\bibinfo{person}{Bita~Darvish Rouhani},
  \bibinfo{person}{M~Sadegh Riazi}, {and} \bibinfo{person}{Farinaz
  Koushanfar}.} \bibinfo{year}{2018}\natexlab{}.
\newblock \showarticletitle{Deepsecure: Scalable provably-secure deep
  learning}. In \bibinfo{booktitle}{\emph{Proceedings of the 55th annual design
  automation conference}}. \bibinfo{pages}{1--6}.
\newblock


\bibitem[Salem et~al\mbox{.}(2020)]%
        {salem2020updates}
\bibfield{author}{\bibinfo{person}{Ahmed Salem}, \bibinfo{person}{Apratim
  Bhattacharya}, \bibinfo{person}{Michael Backes}, \bibinfo{person}{Mario
  Fritz}, {and} \bibinfo{person}{Yang Zhang}.} \bibinfo{year}{2020}\natexlab{}.
\newblock \showarticletitle{$\{$Updates-Leak$\}$: Data set inference and
  reconstruction attacks in online learning}. In \bibinfo{booktitle}{\emph{29th
  USENIX security symposium (USENIX Security 20)}}.
  \bibinfo{pages}{1291--1308}.
\newblock


\bibitem[Sammaknejad et~al\mbox{.}(2017)]%
        {sammaknejad2017gender}
\bibfield{author}{\bibinfo{person}{Negar Sammaknejad},
  \bibinfo{person}{Hamidreza Pouretemad}, \bibinfo{person}{Changiz Eslahchi},
  \bibinfo{person}{Alireza Salahirad}, {and} \bibinfo{person}{Ashkan
  Alinejad}.} \bibinfo{year}{2017}\natexlab{}.
\newblock \showarticletitle{Gender classification based on eye movements: A
  processing effect during passive face viewing}.
\newblock \bibinfo{journal}{\emph{Advances in Cognitive Psychology}}
  \bibinfo{volume}{13}, \bibinfo{number}{3} (\bibinfo{year}{2017}),
  \bibinfo{pages}{232}.
\newblock


\bibitem[Shokri et~al\mbox{.}(2017)]%
        {shokri2017membership}
\bibfield{author}{\bibinfo{person}{Reza Shokri}, \bibinfo{person}{Marco
  Stronati}, \bibinfo{person}{Congzheng Song}, {and} \bibinfo{person}{Vitaly
  Shmatikov}.} \bibinfo{year}{2017}\natexlab{}.
\newblock \showarticletitle{Membership inference attacks against machine
  learning models}. In \bibinfo{booktitle}{\emph{IEEE S\&P}}.
  \bibinfo{pages}{3--18}.
\newblock


\bibitem[Smith et~al\mbox{.}(2013)]%
        {smith2013gaze}
\bibfield{author}{\bibinfo{person}{Brian~A Smith}, \bibinfo{person}{Qi Yin},
  \bibinfo{person}{Steven~K Feiner}, {and} \bibinfo{person}{Shree~K Nayar}.}
  \bibinfo{year}{2013}\natexlab{}.
\newblock \showarticletitle{Gaze locking: passive eye contact detection for
  human-object interaction}. In \bibinfo{booktitle}{\emph{Proceedings of the
  26th annual ACM symposium on User interface software and technology}}.
  \bibinfo{pages}{271--280}.
\newblock


\bibitem[Steil and Bulling(2015)]%
        {steil2015discovery}
\bibfield{author}{\bibinfo{person}{Julian Steil} {and} \bibinfo{person}{Andreas
  Bulling}.} \bibinfo{year}{2015}\natexlab{}.
\newblock \showarticletitle{Discovery of everyday human activities from
  long-term visual behaviour using topic models}. In
  \bibinfo{booktitle}{\emph{Proceedings of the 2015 acm international joint
  conference on pervasive and ubiquitous computing}}. \bibinfo{pages}{75--85}.
\newblock


\bibitem[Steil et~al\mbox{.}(2019)]%
        {steil2019privacy}
\bibfield{author}{\bibinfo{person}{Julian Steil}, \bibinfo{person}{Inken
  Hagestedt}, \bibinfo{person}{Michael~Xuelin Huang}, {and}
  \bibinfo{person}{Andreas Bulling}.} \bibinfo{year}{2019}\natexlab{}.
\newblock \showarticletitle{Privacy-aware eye tracking using differential
  privacy}. In \bibinfo{booktitle}{\emph{ACM Symposium on Eye Tracking Research
  \& Applications}}. \bibinfo{pages}{1--9}.
\newblock


\bibitem[Sugano et~al\mbox{.}(2014)]%
        {sugano2014learning}
\bibfield{author}{\bibinfo{person}{Yusuke Sugano}, \bibinfo{person}{Yasuyuki
  Matsushita}, {and} \bibinfo{person}{Yoichi Sato}.}
  \bibinfo{year}{2014}\natexlab{}.
\newblock \showarticletitle{Learning-by-synthesis for appearance-based 3d gaze
  estimation}. In \bibinfo{booktitle}{\emph{Proceedings of the IEEE conference
  on computer vision and pattern recognition}}. \bibinfo{pages}{1821--1828}.
\newblock


\bibitem[Tonsen et~al\mbox{.}(2017)]%
        {tonsen2017invisible}
\bibfield{author}{\bibinfo{person}{Marc Tonsen}, \bibinfo{person}{Julian
  Steil}, \bibinfo{person}{Yusuke Sugano}, {and} \bibinfo{person}{Andreas
  Bulling}.} \bibinfo{year}{2017}\natexlab{}.
\newblock \showarticletitle{Invisibleeye: Mobile eye tracking using multiple
  low-resolution cameras and learning-based gaze estimation}.
\newblock \bibinfo{journal}{\emph{Proceedings of the ACM on Interactive,
  Mobile, Wearable and Ubiquitous Technologies}} \bibinfo{volume}{1},
  \bibinfo{number}{3} (\bibinfo{year}{2017}), \bibinfo{pages}{1--21}.
\newblock


\bibitem[Tonsen et~al\mbox{.}(2016)]%
        {tonsen2016lpw}
\bibfield{author}{\bibinfo{person}{Marc Tonsen}, \bibinfo{person}{Xucong
  Zhang}, \bibinfo{person}{Yusuke Sugano}, {and} \bibinfo{person}{Andreas
  Bulling}.} \bibinfo{year}{2016}\natexlab{}.
\newblock \showarticletitle{Labelled Pupils in the Wild: A Dataset for Studying
  Pupil Detection in Unconstrained Environments}. In
  \bibinfo{booktitle}{\emph{Proceedings of the Ninth Biennial ACM Symposium on
  Eye Tracking Research \& Applications}} (Charleston, South Carolina)
  \emph{(\bibinfo{series}{ETRA '16})}. \bibinfo{publisher}{Association for
  Computing Machinery}, \bibinfo{address}{New York, NY, USA},
  \bibinfo{pages}{139–142}.
\newblock
\showISBNx{9781450341257}
\urldef\tempurl%
\url{https://doi.org/10.1145/2857491.2857520}
\showDOI{\tempurl}


\bibitem[Valenti et~al\mbox{.}(2011)]%
        {valenti2011combining}
\bibfield{author}{\bibinfo{person}{Roberto Valenti}, \bibinfo{person}{Nicu
  Sebe}, {and} \bibinfo{person}{Theo Gevers}.} \bibinfo{year}{2011}\natexlab{}.
\newblock \showarticletitle{Combining head pose and eye location information
  for gaze estimation}.
\newblock \bibinfo{journal}{\emph{IEEE Transactions on Image Processing}}
  \bibinfo{volume}{21}, \bibinfo{number}{2} (\bibinfo{year}{2011}),
  \bibinfo{pages}{802--815}.
\newblock


\bibitem[Vertegaal et~al\mbox{.}(2003)]%
        {vertegaal2003attentive}
\bibfield{author}{\bibinfo{person}{Roel Vertegaal} {et~al\mbox{.}}}
  \bibinfo{year}{2003}\natexlab{}.
\newblock \showarticletitle{Attentive user interfaces}.
\newblock \bibinfo{journal}{\emph{Commun. ACM}} \bibinfo{volume}{46},
  \bibinfo{number}{3} (\bibinfo{year}{2003}), \bibinfo{pages}{30--33}.
\newblock


\bibitem[Westerlund(2019)]%
        {westerlund2019emergence}
\bibfield{author}{\bibinfo{person}{Mika Westerlund}.}
  \bibinfo{year}{2019}\natexlab{}.
\newblock \showarticletitle{The emergence of deepfake technology: A review}.
\newblock \bibinfo{journal}{\emph{Technology innovation management review}}
  \bibinfo{volume}{9}, \bibinfo{number}{11} (\bibinfo{year}{2019}).
\newblock


\bibitem[Wood et~al\mbox{.}(2016a)]%
        {wood16_eccv}
\bibfield{author}{\bibinfo{person}{Erroll Wood}, \bibinfo{person}{Tadas
  Baltru{\v{s}}aitis}, \bibinfo{person}{Louis-Philippe Morency},
  \bibinfo{person}{Peter Robinson}, {and} \bibinfo{person}{Andreas Bulling}.}
  \bibinfo{year}{2016}\natexlab{a}.
\newblock \showarticletitle{A 3D Morphable Eye Region Model for Gaze
  Estimation}. In \bibinfo{booktitle}{\emph{Proc. European Conference on
  Computer Vision (ECCV)}}. \bibinfo{pages}{297--313}.
\newblock
\urldef\tempurl%
\url{https://doi.org/10.1007/978-3-319-46448-0_18}
\showDOI{\tempurl}


\bibitem[Wood et~al\mbox{.}(2016b)]%
        {wood16_etra}
\bibfield{author}{\bibinfo{person}{Erroll Wood}, \bibinfo{person}{Tadas
  Baltru{\v{s}}aitis}, \bibinfo{person}{Louis-Philippe Morency},
  \bibinfo{person}{Peter Robinson}, {and} \bibinfo{person}{Andreas Bulling}.}
  \bibinfo{year}{2016}\natexlab{b}.
\newblock \showarticletitle{Learning an appearance-based gaze estimator from
  one million synthesised images}. In \bibinfo{booktitle}{\emph{Proc. ACM
  International Symposium on Eye Tracking Research and Applications (ETRA)}}.
  \bibinfo{pages}{131--138}.
\newblock
\urldef\tempurl%
\url{https://doi.org/10.1145/2857491.2857492}
\showDOI{\tempurl}


\bibitem[Wu et~al\mbox{.}(2016)]%
        {wu2016methodology}
\bibfield{author}{\bibinfo{person}{Xi Wu}, \bibinfo{person}{Matthew
  Fredrikson}, \bibinfo{person}{Somesh Jha}, {and} \bibinfo{person}{Jeffrey~F
  Naughton}.} \bibinfo{year}{2016}\natexlab{}.
\newblock \showarticletitle{A methodology for formalizing model-inversion
  attacks}. In \bibinfo{booktitle}{\emph{2016 IEEE 29th Computer Security
  Foundations Symposium (CSF)}}. IEEE, \bibinfo{pages}{355--370}.
\newblock


\bibitem[Yang et~al\mbox{.}(2019)]%
        {yang2019diversity}
\bibfield{author}{\bibinfo{person}{Dingdong Yang}, \bibinfo{person}{Seunghoon
  Hong}, \bibinfo{person}{Yunseok Jang}, \bibinfo{person}{Tianchen Zhao}, {and}
  \bibinfo{person}{Honglak Lee}.} \bibinfo{year}{2019}\natexlab{}.
\newblock \showarticletitle{Diversity-sensitive conditional generative
  adversarial networks}.
\newblock \bibinfo{journal}{\emph{arXiv preprint arXiv:1901.09024}}
  (\bibinfo{year}{2019}).
\newblock


\bibitem[Yu et~al\mbox{.}(2017)]%
        {yu17user}
\bibfield{author}{\bibinfo{person}{Kun Yu}, \bibinfo{person}{Shlomo Berkovsky},
  \bibinfo{person}{Ronnie Taib}, \bibinfo{person}{Dan Conway},
  \bibinfo{person}{Jianlong Zhou}, {and} \bibinfo{person}{Fang Chen}.}
  \bibinfo{year}{2017}\natexlab{}.
\newblock \showarticletitle{User Trust Dynamics: An Investigation Driven by
  Differences in System Performance}. In \bibinfo{booktitle}{\emph{Proceedings
  of the 22nd International Conference on Intelligent User Interfaces}}
  (Limassol, Cyprus) \emph{(\bibinfo{series}{IUI '17})}.
  \bibinfo{publisher}{Association for Computing Machinery},
  \bibinfo{address}{New York, NY, USA}, \bibinfo{pages}{307–317}.
\newblock
\showISBNx{9781450343480}
\urldef\tempurl%
\url{https://doi.org/10.1145/3025171.3025219}
\showDOI{\tempurl}


\bibitem[Yu and Odobez(2020)]%
        {yu2020unsupervised}
\bibfield{author}{\bibinfo{person}{Yu Yu} {and} \bibinfo{person}{Jean-Marc
  Odobez}.} \bibinfo{year}{2020}\natexlab{}.
\newblock \showarticletitle{Unsupervised representation learning for gaze
  estimation}. In \bibinfo{booktitle}{\emph{Proceedings of the IEEE/CVF
  Conference on Computer Vision and Pattern Recognition}}.
  \bibinfo{pages}{7314--7324}.
\newblock


\bibitem[Zhang et~al\mbox{.}(2018)]%
        {zhang18_chi}
\bibfield{author}{\bibinfo{person}{Xucong Zhang},
  \bibinfo{person}{Michael~Xuelin Huang}, \bibinfo{person}{Yusuke Sugano},
  {and} \bibinfo{person}{Andreas Bulling}.} \bibinfo{year}{2018}\natexlab{}.
\newblock \showarticletitle{Training Person-Specific Gaze Estimators from
  Interactions with Multiple Devices}. In \bibinfo{booktitle}{\emph{Proc. ACM
  SIGCHI Conference on Human Factors in Computing Systems (CHI)}}.
  \bibinfo{pages}{1--12}.
\newblock
\urldef\tempurl%
\url{https://doi.org/10.1145/3173574.3174198}
\showDOI{\tempurl}


\bibitem[Zhang et~al\mbox{.}(2020b)]%
        {zhang2020eth}
\bibfield{author}{\bibinfo{person}{Xucong Zhang}, \bibinfo{person}{Seonwook
  Park}, \bibinfo{person}{Thabo Beeler}, \bibinfo{person}{Derek Bradley},
  \bibinfo{person}{Siyu Tang}, {and} \bibinfo{person}{Otmar Hilliges}.}
  \bibinfo{year}{2020}\natexlab{b}.
\newblock \showarticletitle{Eth-xgaze: A large scale dataset for gaze
  estimation under extreme head pose and gaze variation}. In
  \bibinfo{booktitle}{\emph{European Conference on Computer Vision}}. Springer,
  \bibinfo{pages}{365--381}.
\newblock


\bibitem[Zhang et~al\mbox{.}(2017b)]%
        {zhang2017everyday}
\bibfield{author}{\bibinfo{person}{Xucong Zhang}, \bibinfo{person}{Yusuke
  Sugano}, {and} \bibinfo{person}{Andreas Bulling}.}
  \bibinfo{year}{2017}\natexlab{b}.
\newblock \showarticletitle{Everyday eye contact detection using unsupervised
  gaze target discovery}. In \bibinfo{booktitle}{\emph{Proceedings of the 30th
  annual ACM symposium on user interface software and technology}}.
  \bibinfo{pages}{193--203}.
\newblock


\bibitem[Zhang et~al\mbox{.}(2015)]%
        {zhang15_cvpr}
\bibfield{author}{\bibinfo{person}{Xucong Zhang}, \bibinfo{person}{Yusuke
  Sugano}, \bibinfo{person}{Mario Fritz}, {and} \bibinfo{person}{Andreas
  Bulling}.} \bibinfo{year}{2015}\natexlab{}.
\newblock \showarticletitle{Appearance-based Gaze Estimation in the Wild}. In
  \bibinfo{booktitle}{\emph{Proc. IEEE Conference on Computer Vision and
  Pattern Recognition (CVPR)}}. \bibinfo{pages}{4511--4520}.
\newblock
\urldef\tempurl%
\url{https://doi.org/10.1109/CVPR.2015.7299081}
\showDOI{\tempurl}


\bibitem[Zhang et~al\mbox{.}(2017c)]%
        {zhang2017s}
\bibfield{author}{\bibinfo{person}{Xucong Zhang}, \bibinfo{person}{Yusuke
  Sugano}, \bibinfo{person}{Mario Fritz}, {and} \bibinfo{person}{Andreas
  Bulling}.} \bibinfo{year}{2017}\natexlab{c}.
\newblock \showarticletitle{It's written all over your face: Full-face
  appearance-based gaze estimation}. In \bibinfo{booktitle}{\emph{Proceedings
  of the IEEE conference on computer vision and pattern recognition
  workshops}}. \bibinfo{pages}{51--60}.
\newblock


\bibitem[Zhang et~al\mbox{.}(2019)]%
        {zhang19_pami}
\bibfield{author}{\bibinfo{person}{Xucong Zhang}, \bibinfo{person}{Yusuke
  Sugano}, \bibinfo{person}{Mario Fritz}, {and} \bibinfo{person}{Andreas
  Bulling}.} \bibinfo{year}{2019}\natexlab{}.
\newblock \showarticletitle{MPIIGaze: Real-World Dataset and Deep
  Appearance-Based Gaze Estimation}.
\newblock \bibinfo{journal}{\emph{IEEE Transactions on Pattern Analysis and
  Machine Intelligence (TPAMI)}} \bibinfo{volume}{41}, \bibinfo{number}{1}
  (\bibinfo{year}{2019}), \bibinfo{pages}{162--175}.
\newblock
\urldef\tempurl%
\url{https://doi.org/10.1109/TPAMI.2017.2778103}
\showDOI{\tempurl}


\bibitem[Zhang et~al\mbox{.}(2020a)]%
        {zhang2020secret}
\bibfield{author}{\bibinfo{person}{Yuheng Zhang}, \bibinfo{person}{Ruoxi Jia},
  \bibinfo{person}{Hengzhi Pei}, \bibinfo{person}{Wenxiao Wang},
  \bibinfo{person}{Bo Li}, {and} \bibinfo{person}{Dawn Song}.}
  \bibinfo{year}{2020}\natexlab{a}.
\newblock \showarticletitle{The secret revealer: Generative model-inversion
  attacks against deep neural networks}. In
  \bibinfo{booktitle}{\emph{Proceedings of the IEEE/CVF conference on computer
  vision and pattern recognition}}. \bibinfo{pages}{253--261}.
\newblock


\bibitem[Zhang et~al\mbox{.}(2017a)]%
        {zhang2017age}
\bibfield{author}{\bibinfo{person}{Zhifei Zhang}, \bibinfo{person}{Yang Song},
  {and} \bibinfo{person}{Hairong Qi}.} \bibinfo{year}{2017}\natexlab{a}.
\newblock \showarticletitle{Age progression/regression by conditional
  adversarial autoencoder}. In \bibinfo{booktitle}{\emph{Proceedings of the
  IEEE conference on computer vision and pattern recognition}}.
  \bibinfo{pages}{5810--5818}.
\newblock


\bibitem[Zhao et~al\mbox{.}(2020)]%
        {zhao2020idlg}
\bibfield{author}{\bibinfo{person}{Bo Zhao}, \bibinfo{person}{Konda~Reddy
  Mopuri}, {and} \bibinfo{person}{Hakan Bilen}.}
  \bibinfo{year}{2020}\natexlab{}.
\newblock \showarticletitle{idlg: Improved deep leakage from gradients}.
\newblock \bibinfo{journal}{\emph{arXiv preprint arXiv:2001.02610}}
  (\bibinfo{year}{2020}).
\newblock


\end{thebibliography}

\newpage

\section*{Appendix}
  \appendix
\section{Qualitative Evaluation} \label{app:study}
\begin{figure}[H]
\centering
  \includegraphics[width=11cm]{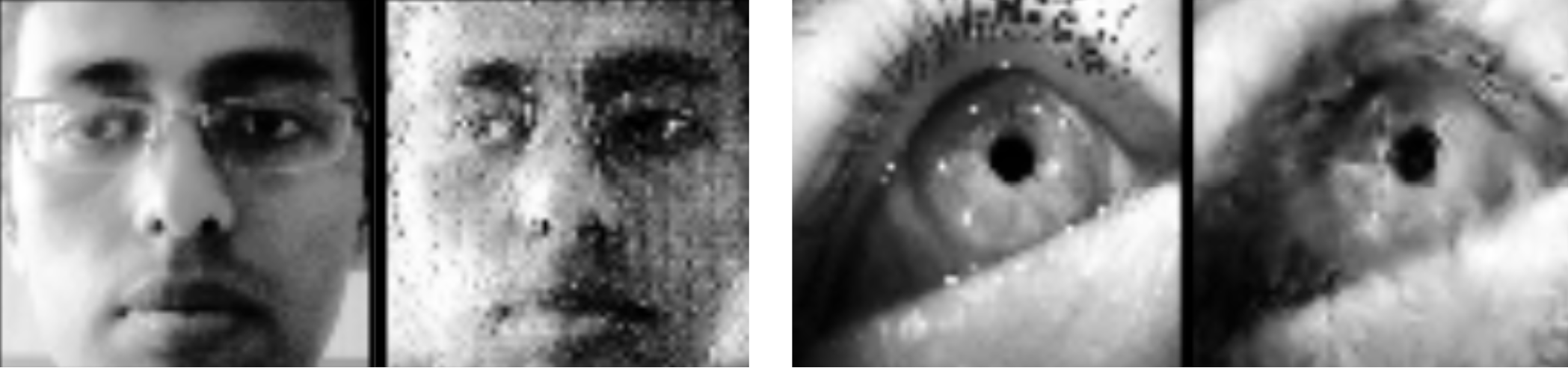}
    \caption{For similar ground truth (left) and reconstructed images (right) from the 3 different datasets, participants were asked the following questions:\\
    A.1.  Do both images represent the same person? (Yes/No)\\
    A.2. On a scale from 1 to 10,  how would you rate the overall similarity of both images? (Very similar to very dissimilar)
} 
\label{fig:study}
\end{figure}

\begin{figure}[H]
\centering
  \includegraphics[width=15cm]{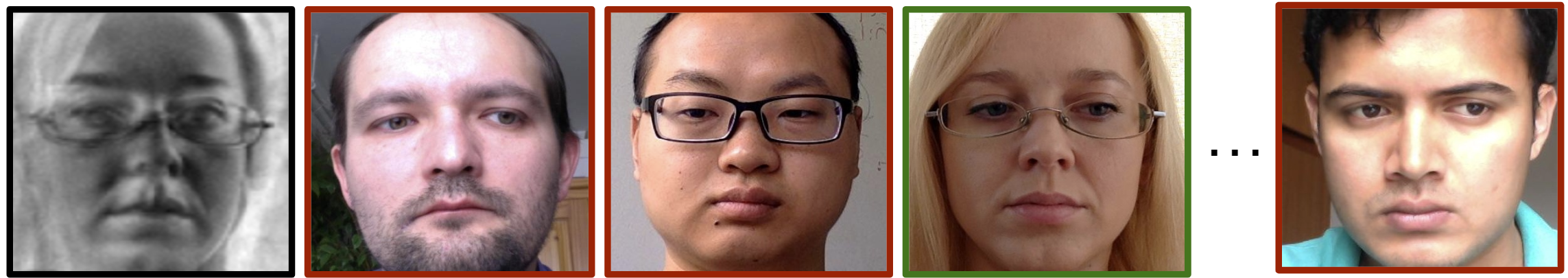}
    \caption{For similar reconstructed images (left), participants were asked to map the full-face/eye image to the corresponding ground-truth clients.} 
\label{fig:study2}
\end{figure}

To get the qualitative assessment of the \attackName~ reconstruction results, a user study was conducted (N = 60 - each image was shown to 3 participants) on a randomly-picked subset of 120 
images covering all cases discussed in \cref{sec:experiments}. Participants were asked to assess each reconstructed image in terms of similarity and re-identification. An example of the survey is shown in \cref{fig:study} and \cref{fig:study2}.

\section{Additional results}\label{sec:additional}
In \cref{tab:attack2} and \cref{tab:attack3}, we show the \attackName~ performance on the remaining datasets, namely MPIIGaze and NVGaze.
The mean angular error (MAE) has been calculated with gaze estimators that already have MAE of 6.3, 6.2, and 0.8 for MPIIGaze, MPIIFaceGaze, and NVGaze, respectively. 

\begin{table}[H]
    \centering

    \begin{tabularx}{1\textwidth} 
    { 
       >{\raggedright \arraybackslash}X 
       >{\centering\arraybackslash}X
       >{\centering\arraybackslash}X
       >{\centering\arraybackslash}X
       >{\centering\arraybackslash}X
       >{\centering\arraybackslash}X
      }
     \hline
      
       & \textbf{Re-identified} &\textbf{Visual score} & \textbf{PSNR} & \textbf{MAE} & \textbf{KL} \\
     \hline
     \multirow{1}{5em}{\textbf{Data centre} }  & 15/15 & 100\% &  max. & 0 & 0\\
     \hline
     \textbf{Adaptive FL}  & 15/15 & 80\% & 18.3 & 7.8 & 0.198 \\
    \hline 
    \textbf\methodName & 0/15  & 4\% & 4.8 & 10.1 & 0.255 \\
     \hline
 \multirow{1}{6em}{\textbf{Generic MPC}}  & 0/15  & 5\% & 4.8 & 9.9 & 0.257 \\ 

\hline       
    \end{tabularx}
    
    \caption{\attackName~performance on the MPIIGaze dataset.
    }
    \label{tab:attack2}
\end{table}

\begin{table}[H]
    \centering

    \begin{tabularx}{1\textwidth} 
    { 
       >{\raggedright \arraybackslash}X 
       >{\centering\arraybackslash}X
       >{\centering\arraybackslash}X
       >{\centering\arraybackslash}X
       >{\centering\arraybackslash}X
       >{\centering\arraybackslash}X
      }
     \hline
      
       & \textbf{Re-identified} &\textbf{Visual score} & \textbf{PSNR} & \textbf{MAE} & \textbf{KL} \\
     \hline
     \multirow{1}{5em}{\textbf{Data centre} }  & 32/32 & 100\% &  max. & 0 & 0\\
     \hline
     \textbf{Adaptive FL}  & 32/32 & 92\% & 12.4 & 2.0 & 0.167 \\
    \hline 
    \textbf\methodName & 0/32  & 13\% & 4.1 & 6.7 & 0.181 \\
     \hline
 \multirow{1}{6em}{\textbf{Generic MPC}}  & 0/32  & 10\% & 4.0 & 6.6 & 0.180 \\ 

\hline       
    \end{tabularx}
    
    \caption{\attackName~performance on the NVGaze dataset.
    }
    \label{tab:attack3}
\end{table}

\begin{figure}[H]
\centering
  \includegraphics[width=10cm]{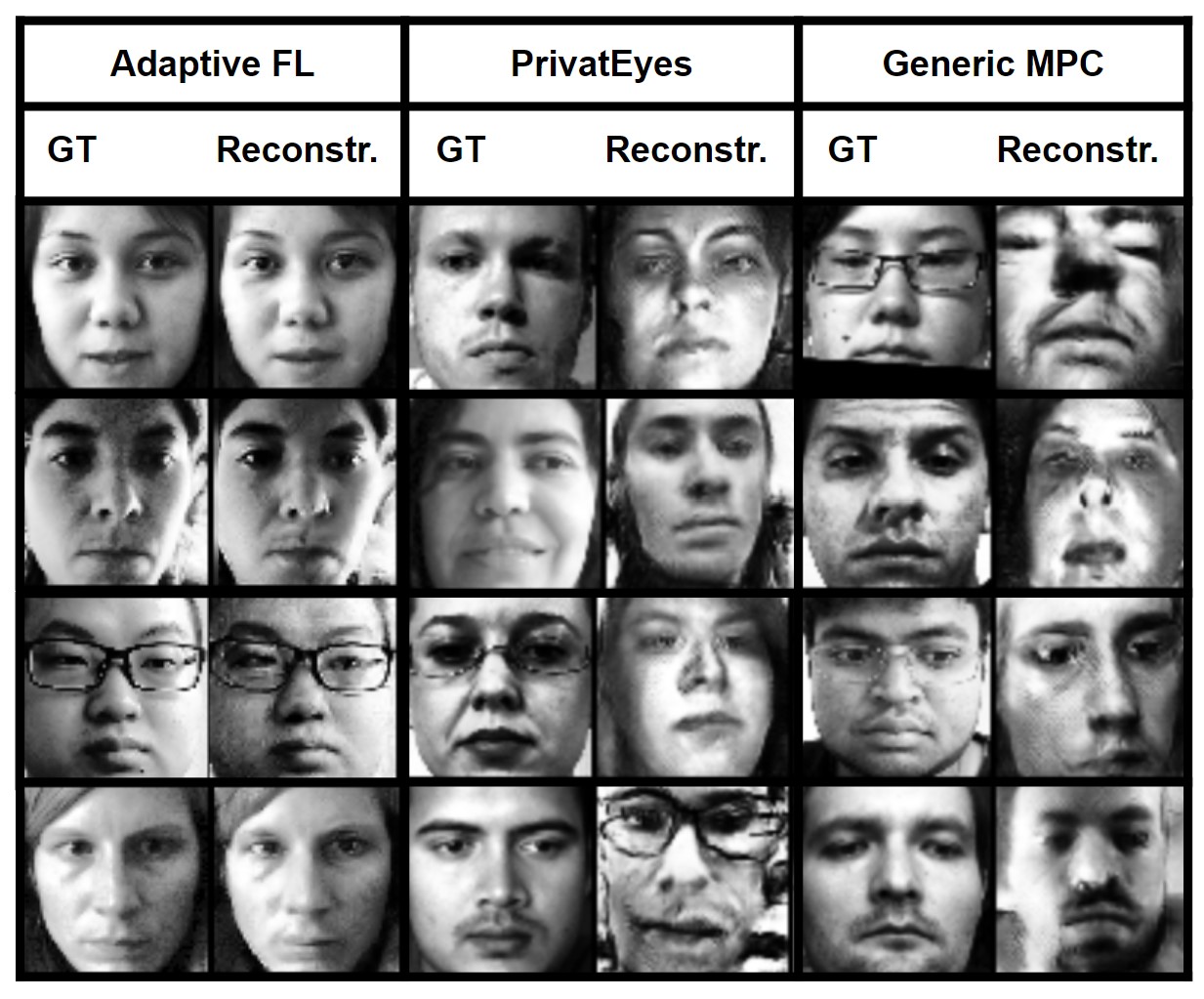}
    \caption{Additional examples of faces reconstructed by \attackName~from the MPIIFaceGaze dataset.} 
\label{fig:additional}
\end{figure}

\begin{figure}[H]
\centering
  \includegraphics[width=10cm]{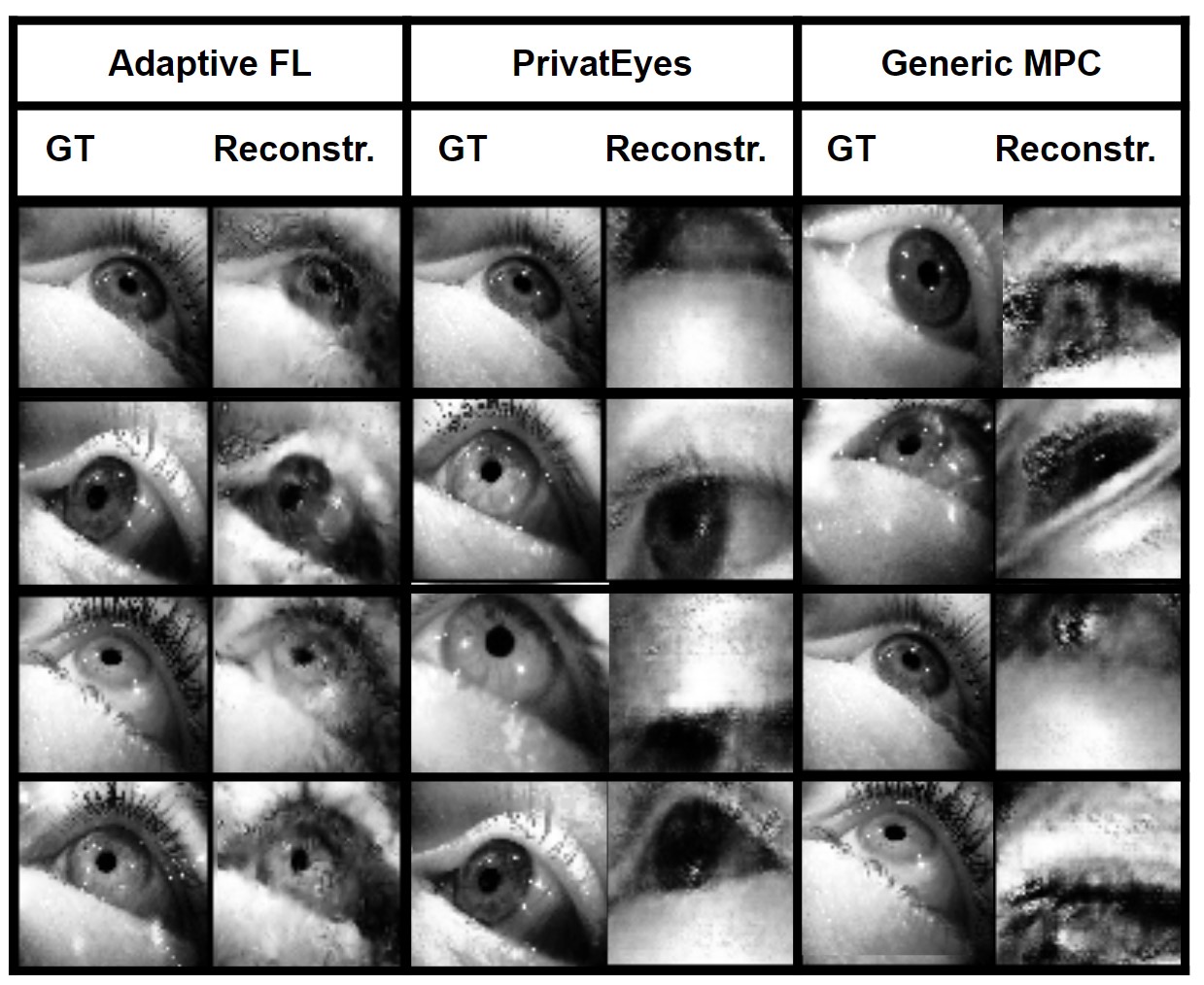}
    \caption{Additional examples of eyes reconstructed by \attackName~from the NVGaze dataset.} 
\label{fig:additional2}
\end{figure}

In \cref{fig:additional} and \cref{fig:additional2}, we show additional examples for reconstructed face and eye images, respectively, covering different participants (i.e. appearances) and gaze directions. As previously proven by our evaluation metrics (c.f. \cref{sec:experiments}), \methodName~and generic MPC leak less information in comparison to adaptive FL. The resulting faces/eyes remain nonetheless realistic due to the discriminator loss in \attackName.

\begin{figure}[H]
\centering
\includegraphics[width=10cm]{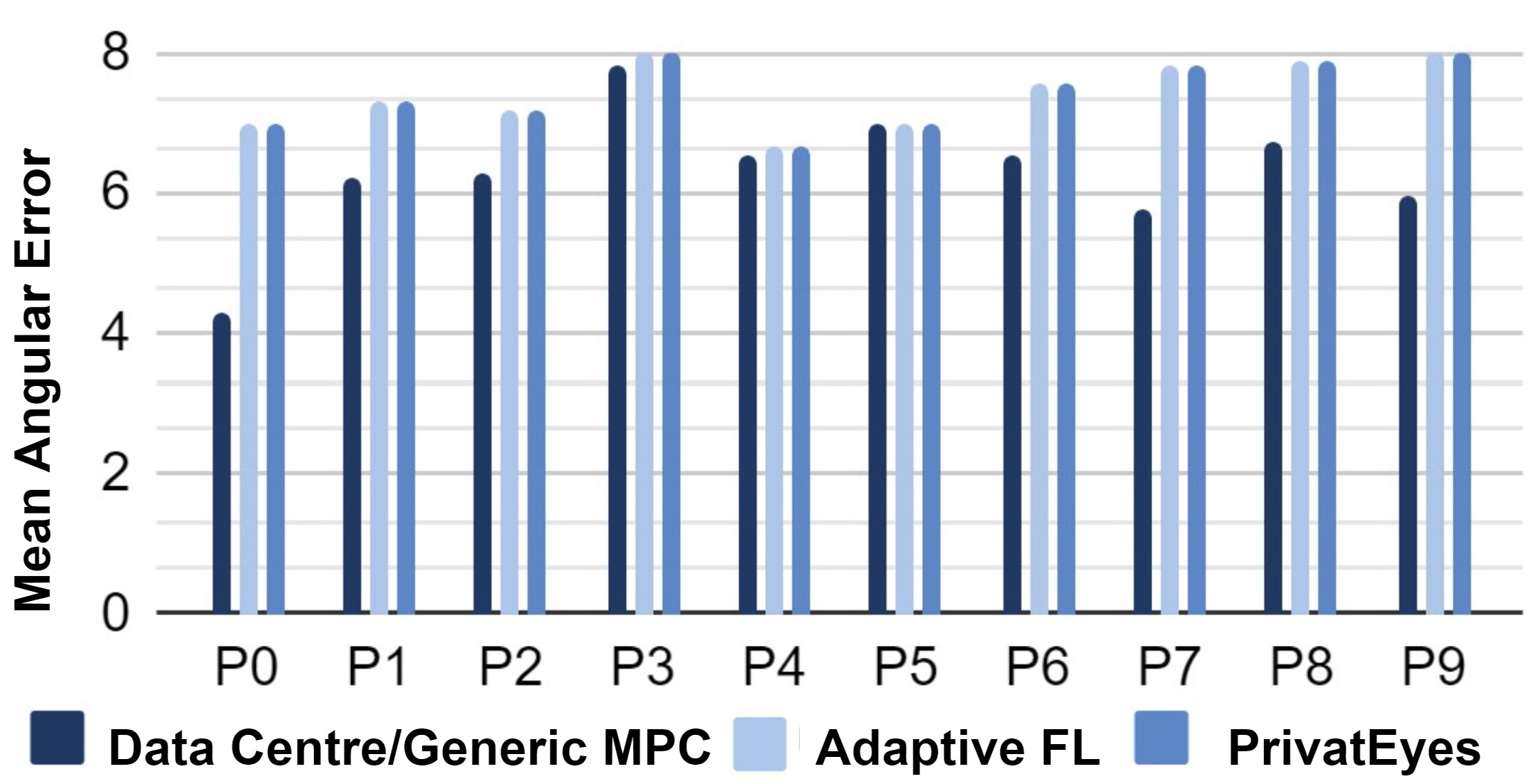}
\caption{Mean gaze estimation error for 10 randomly-selected participants from MPIIFaceGaze. 
}
\label{fig:individual}
\end{figure}

    

\section{Convergence effect on privacy} \label{app:convergence}

\begin{figure}[H]
\centering
  \includegraphics[width=15cm]{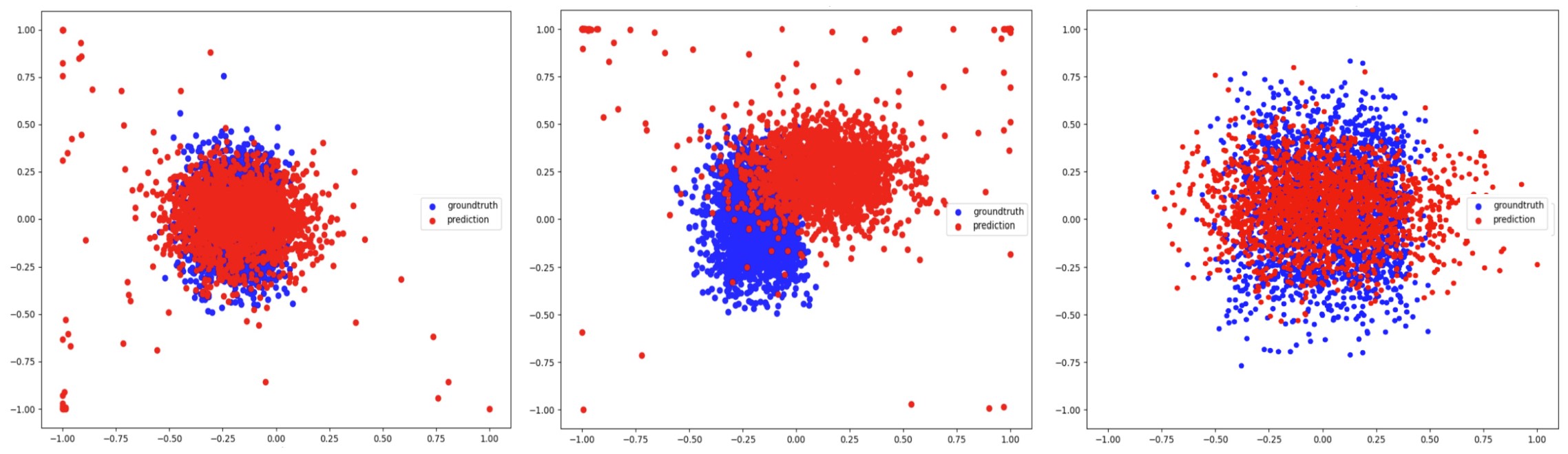}
    \caption{Example of the ground-truth vs predicted gaze data distribution of the normalised gaze directions with KL-divergence = 0.198 (left), KL-divergence = 0.260 (middle), and KL-divergence = 0.202 (right).} 
\label{fig:dist}
\end{figure}

In \cref{fig:dist}, the convergence effect can be seen from the left figure (i.e. the first training round) to the middle figures (i.e. a later training round) where some data samples no longer contribute to updating the trained model (gradients). 
Without aux. knowledge (middle), the prediction takes into account all training samples that previously contributed to the training of the entire model regardless of the current data of the target model. Hence, adding the previous model updates as aux. knowledge (right) enhances the reconstruction (i.e. KL-divergence).

\section{Robustness to other data leakage attacks} \label{sub:attacks}
We further investigated the potential information leakage by other existing generic attacks using the ML-doctor \cite{liu2022mldoc} framework which benchmarks state-of-the-art attacks on ML models.

\noindentparagraph{Membership inference attacks}
Membership inference attacks \cite{shokri2017membership} aim to determine if a particular data sample was used to train the model. For instance, in privacy-sensitive gaze applications (e.g. mental disorders detection \cite{paletta2020mira, higuchi2018visualizing, li2022appearance}), this membership knowledge could imply the inclusion of an individual in a certain group (e.g. mental disorder).
To perform the attack the adversary constructs a shadow dataset to train a shadow model 
on the same task (gaze estimation). 
In addition, the adversary trains a binary classifier (member or non-member) by querying the shadow model on training and out-of-training shadow datasets. 
Finally, target images are fed to the target model (the gaze estimator) and posteriors are compared to the binary classifier.
For all datasets, the attack correctly identified more than 90\%(first round) to 74\% (last round) of the clients from individual updates (adaptive FL).
For output models (\methodName), for the MPIIFaceGaze dataset, membership inference was less successful as the number of rounds increased, i.e. the average accuracy dropped from 40\% (first round) to 33\% (last round) correctly determined samples. Moreover, for the MPIIGaze dataset, the attack's accuracy dropped from 33\% (first round) to 13\% (last round).
Similarly, for the NVGaze dataset, the attack's accuracy dropped from 37\% (first round) to 12\% (last round).
Therefore, \methodName~reduces the gaze information leakage for membership inference attacks significantly due to the inaccessibility of individual updates and the information obfuscation during the aggregation step.

\noindentparagraph{Attribute inference attacks}

Attribute inference attacks \cite{melis2019exploiting} exploit additional knowledge gained by ML models, which is not needed for the training task, to infer private attributes about the data. This attack is similar to membership inference attacks but substitutes the binary classifier with the attack model trained on the to-be-inferred attributes\footnote{In \attackName, the adversary reconstructs the input image, the gaze distributions, and the corresponding loss values which could later be fed to attribute classifiers.}. 
We opted for the simple attack of predicting the gender of the participants whose data was used to train the gaze estimation model.
Consequently, the attack model is pre-trained on the UTKFace dataset \cite{zhang2017age}.
For all datasets, the attack correctly predicted more than 95\% (first round) to 65\% (last round) of clients' genders from individual updates.
From output models, the attack correctly predicted the gender of 5/15 (first round) to 2/15 (last round) participants for MPIIFaceGaze, 3/15 to 0/15 for MPIIGaze, and 10/32 to 1/32 for NVGaze.


\end{document}